\documentclass[pdflatex,sn-basic]{sn-jnl}

\usepackage{graphicx}%
\usepackage{multirow}%
\usepackage{amsmath,amssymb,amsfonts}%
\usepackage{amsthm}%
\usepackage{mathrsfs}%
\usepackage[title]{appendix}%
\usepackage{xcolor, colortbl}%
\usepackage{textcomp}%
\usepackage{manyfoot}%
\usepackage{booktabs}%
\usepackage{algorithm}%
\usepackage{algorithmicx}%
\usepackage{algpseudocode}%
\usepackage{listings}%
\usepackage{subfigure}
\usepackage{rotating}
\usepackage{wrapfig,lipsum}
\usepackage{tcolorbox}

\definecolor{CherryBlossomPink}{rgb}{1.0, 0.72, 0.77}
\definecolor{applegreen}{rgb}{0.55, 0.71, 0.0}
\definecolor{candyapplered}{rgb}{1.0, 0.03, 0.0}

\theoremstyle{thmstyleone}%
\theoremstyle{thmstyletwo}%

\theoremstyle{thmstylethree}%

\begin{document}

\title[Article Title]{Uncovering implementable dormant pruning decisions from three different stakeholder perspectives}


\author*[1]{\fnm{Deanna} \sur{Flynn}}\email{flynnde@oregonstate.edu}

\author[1]{\fnm{Abhinav} \sur{Jain}}\email{jainab@oregonstate.edu}
\author[1]{\fnm{Heather} \sur{Knight}}\email{knighth@oregonstate.edu}
\author[1]{\fnm{Cristina G.} \sur{Wilson}}\email{wilsoncr@oregonstate.edu}

\author[1]{\fnm{Cindy} \sur{Grimm}} \email{grimmc@oregonstate.edu}

\affil[1]{\orgdiv{Collaborative Robotics and Intelligent Systems Institute}, \orgname{Oregon State University}, \orgaddress{\street{P.O. Box 1212}, \city{Corvallis}, \postcode{43017-6221}, \state{Oregon}, \country{USA}}}




\abstract{Dormant pruning, or the removal of unproductive portions of a tree while a tree is not actively growing, is an important orchard task to help maintain yield, requiring years to build expertise. Because of long training periods and an increasing labor shortage in agricultural jobs, pruning could benefit from robotic automation. These efforts need not replace skilled human labor, but may instead reduce the overall need for human labor by letting robot pruners do significant starting cuts before human pruners finish the job. However, to program robots to prune branches, we first need to understand how pruning decisions are made, and what variables in the environment (e.g., branch size and thickness) we need to capture. Working directly with {\em three} pruning stakeholders --- horticulturists, growers, and pruners --- we find that each group of human experts approaches pruning decision-making differently. To capture this knowledge, we present three studies and two extracted pruning protocols from field work conducted in Prosser, Washington in January 2022 and 2023. We interviewed six stakeholders (two in each group) and observed pruning across three cultivars --- Bing Cherries, Envy Apples, and Jazz Apples --- and two tree architectures --- Upright Fruiting Offshoot and V-Trellis. Leveraging participant interviews and video data, this analysis uses grounded coding to extract pruning terminology, discover horticultural contexts that influence pruning decisions, and find implementable pruning heuristics for autonomous systems. The results include a validated terminology set, which we offer for use by both pruning stakeholders and roboticists, to communicate general pruning concepts and heuristics. The results also highlight seven pruning heuristics utilizing this terminology set that would be relevant for use by future autonomous robot pruning systems, and characterize three discovered horticultural contexts (i.e., environmental management, crop-load management, and replacement wood) across all three cultivars.}




\maketitle

\section{Introduction}
Dormant pruning in orchards consist of the removal of unproductive portions of a tree or plant to maximize fruit production and growth stability over time. Pruning encourages new growth, manages sunlight distribution throughout a tree, and controls fruit productivity, quality, and placement (\citet{alridiwirsah2020effect,carroll2017annual,kolmanivc2021algorithm}). The time and skill required to prune makes dormant pruning the second-most labor intensive and time-consuming agricultural jobs in orchards, typically lasting from mid-November to mid-April (\citet{nae2019immigrants,he2018sensing,gallardo2012cost}). The ability of growers to hire and train enough workers to prune has worsened in recent years given a 75\% decrease in immigration farm labor compared to an 8\% increase in farm labor wages (\citet{nae2019immigrants,agamerica2022labor,daniels2018strawberries}). This labor shortage has led to an interest in automating dormant pruning, but researchers must overcome several barriers before a robotic pruning system can be effectively deployed.

The primary challenge to automated pruning, besides mechanical design and perception, is how to make the decision of where to cut. Pruning is plant and cultivar-specific and depends on many variables. Example variables include: the type and age of cultivar, previous growth and crop-yield behavior, tree architecture (e.g., V-trellis or vertical wall), the current and potential environmental conditions, and desired crop-load and quality. How pruners weigh these variables to produce cut decisions for a specific branch is unknown. Moreover, these pruning decisions --— in practice --— are intuitive, relying on visual and spatial heuristics learned from repetition and experiential learning. Therefore, automating pruning decisions will require making these intuitive and implicit decision processes explicit and algorithmically representable. 

\begin{figure}
    \centering
    \includegraphics[width=0.85\linewidth]{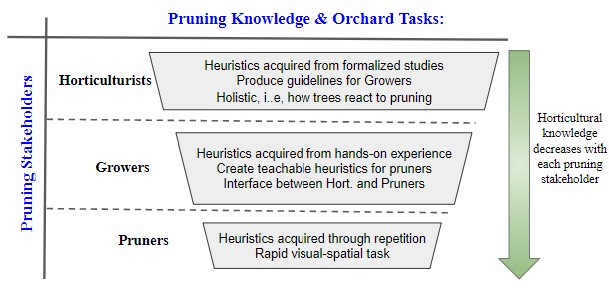}
    \caption{Dormant pruning has three major stakeholders: horticulturists, growers, and pruners. Each stakeholder interacts with orchards in a different way (formal studies, crop management, paid per-cut work) and possesses unique pruning knowledge acquired by formal techniques, hands-on experience, or repetition.}
    \label{fig:stakeholders}
\end{figure}

As part of the study presented in this paper, our analysis has identified {\bf three key stakeholders} --— horticulturists, growers, and pruners —-- each of whom are responsible for different parts of the pruning decision-making process (see Figure~\ref{fig:stakeholders}). It is not possible to fully automate pruning without understanding all three perspectives, how each stakeholder acquires their pruning knowledge, and how to communicate with each stakeholder. Horticulturists have a holistic understanding acquired through formalized studies that look at how particular actions affect the tree over the seasons. For example, they might study how a tree’s cycle reacts to particular pruning cuts (e.g.,~\citet{kupka2007growth,persello2019nature}), how the overall flow of nitrogen throughout the tree affects growth (\citet{kaba2021new}), or how the tree progresses from bloom to blossom to fruit (\citet{zhang2022different}). In comparison, growers (or crew managers) have more hands-on, experience-based knowledge about the link between pruning and tree growth response, gathered from years of experience with large orchards. Growers also interface between the horticulturists and pruners. They distill their in-field experience, plus information provided by horticulturalists, into heuristics that are doable and fast to teach, but still yield quality fruit production with minimal labor. Finally, the actual pruners focus on performing physical cuts as fast as possible (about 1 to 2 cuts per second) since they are often paid by tree or have target pruning goals set by growers.

To better disentangle the complex interplay between these stakeholders and pruning decisions, we conducted three studies with six pruning stakeholders. From these studies we extracted pruning terminology, decisions that affect pruning, and heuristics across three different tree cultivars --- Bing Cherries, Envy Apples, and Jazz Apples --- and two architectures (UFO, V-trellis). The first study was an in-field, in-person, formative study conducted in January 2022. It consisted of two structured stakeholder interviews along with two in-field pruning protocols. These protocols involved observing and interacting with people who were pruning Bing Cherries (UFO) and Envy Apples (V-trellis). The second study was an informal grower interview discussing example pruning decisions made on 28 Jazz Apple trees (V-trellis) that were manually marked with every cut labeled by type. Finally, our third study was a two-hour virtual, discussion-based interview validating pruning terminology and heuristics with two horticulturists. 

Our analysis of these three studies is divided into two sections. The first section provides the set of pruning terminology that the stakeholders use to communicate their pruning heuristics. The second section reports the underlying organization we extracted from the interviews; namely, global goals, mid-level heuristics used to implement those goals in practice, and, finally, the actual pruning cuts that arise from those heuristics. This analysis clearly identifies the characteristics of a tree the automated system must be able to identify, and provides a vocabulary and variables that are important for communicating an automated pruning decision {\em back} to the stakeholders (i.e., why the system chose to make the cut it did).

The remainder of this paper is outlined as follows. Section~\ref{sec:related} reports other successful automated pruning systems, their implementation gaps, and documented pruning heuristics utilized in these systems. In section~\ref{sec:study}, we describe the three study designs for extracting and validating pruning terminology and heuristics. Our discovered pruning terminologies are in section~\ref{sec:terminology}, and example heuristics identified for the two different tree architectures (UFO and V-trellis) and three cultivars (Bing cherry, Jazz and Envy apples) in section~\ref{sec:heuristics}. Finally, section~\ref{sec:discussion} discusses how future researchers can utilize this study methodology in other pruning contexts, and how to translate the discovered terminology and heuristics into an automated robotic pruning system.

Our paper has three main contributions:
\begin{enumerate}
    \item An in-person interview protocol that mixes in-the-field interviews with pruning observations in order to ground verbal descriptions with tree morphology. 
    \item A pruning terminology set, grounded in images and interactions with physical trees, for communicating with stakeholders about pruning heuristics and goals. 
    \item Organization of the information into a hierarchy reflecting global goals linked to heuristics linked to physical cuts, in a form amenable to algorithmic implementation. 
\end{enumerate}

\section{Related Work}
\label{sec:related}

Pruning instructions/guidelines differ in their degree of ambiguity. An example of an unambiguous pruning rule is: ``remove any side branch longer than 5 inches''. In contrast, a pruning {\em heuristic} requires more human interpretation, for example, ``remove vigorous growth longer than 4 inches that is shading other branches''.  In the following discussion we use the terms ``rule'' and ``heuristic'' to distinguish between the two.  

The majority of existing pruning information comes in the form of heuristics --- exemplars that visually represent a final horticultural goal state of pruning, as seen in pruning guides and pamphlets (e.g., ~\citet{polomski2019backyardpruning,maughanbackyardpruning}). However, people's ability to use these guides and apply them to their own trees is difficult because of the differences between their tree(s) and the idealized before and after exemplars. Determining which pruning cuts on their own tree will result in a tree that is ``close'' to the exemplar requires expertise. This expertise is developed by learning (through experience) rules and heuristics that transform an existing tree into the desired exemplar. 

Previous literature provide pruning rules and heuristics for limb renewal for various apple cultivars with a Tall Spindle architecture. For example, Robin et al. (2008) report a pruning rule for `McIntosh' or `Honeycrisp' apples as ``removing branches that were upright" when the cultivar is 2-4 years old (\citet{robinson2008tall}). Likewise, Schupp et al. (2017) define a mathematically-based pruning heuristic for `Buckeye Gala' apples by pruning branches in decreasing order of branch circumference size until a desired Limb to Trunk Ratio (LTR) --- the ratio of branch size to branch spacing --- is achieved (\citet{schupp2017method, schupp2017pruningwholetreeseverity}). To use this heuristic, the pruner can decide what order to remove branches that have the same branch diameter to reach the desired LTR.  

To obtain pruning rules and heuristics from experts --- so that they may, in turn, be implemented in autonomous pruning systems --- previous researchers have utilized at least one of two traditional methods. These are i) observing the manual pruning process and identifying cut behaviors by visually comparing trees before and after pruning (e.g., \citet{poni2016mechanical,karkee2014identification}) or ii) discussing with experts about pruning strategies (\citet{you2022semantics}). Once developed, researchers integrate these rules and heuristics (simple and complex) in autonomous pruning systems and test by validating cuts against experts, either qualitatively or quantitatively. Qualitative validation asks experts to label the system's cut decisions as being correct or incorrect (\citet{corbett2012aivines}). Quantitative validation, however, compares the branch removal percentage for one tree or the percent similarity in pruning cuts between the autonomous system and experts (\citet{westling2021lidar}). We utilize both pruning observations and qualitative validation in our study to ensure we do not make any \textit{a priori} assumptions about pruning heuristics for certain tree cultivars and architectures and to ensure the heuristics are feasible for both autonomous systems and people. 

Past autonomous pruning systems with successful implementations of formalized pruning rules and heuristics is dependent on converting these rules and heuristics into implementable algorithms. Implementing ``simple'' pruning rules and heuristics require little to no modification by researchers. For example, one woody structure that has a simple enough pruning rule to develop and automate is pruning grapevines for light distribution. This is a hedging rule --- cut all branches to a specified length relative to the vine's primary cordon/wire support (e.g., \citet{poni2016mechanical,botterill2017robot,sevila1985robot}). However, implementation failures occur when there is a disconnect between the goal-state exemplar and the visual-spatial tasks the machine must perform (i.e., where and how to cut a branch). These failures happen more frequently in orchards with more intricate tree architectures and complex rules (like apple orchards with V-trellis or Tall Spindle architectures). This is due to the fact that the cuts are precise in their placement and rely on multiple tree factors and global objectives (like branch direction or crop-yield expectations (\citet{he2018sensing, akbar2016novel})). On the other hand, implementing ``complex'' pruning rules and heuristics requires converting pruning decisions --- connecting global orchard objectives to individual cuts --- into quantifiable metrics for evaluation. For example, Westling et al (2011) addressed the connection between pruning and light distribution throughout an avocado and mango tree canopy (\citet{westling2021lidar}). Their system used LiDAR tree scans to select and remove branches negatively impacting light distribution. Their method achieved a 25\% improvement in light distribution compared to a 16\% over commercial pruning. Other researchers have explored where to prune based on 3D topological models of grapefruit vines (\citet{corbett2012aivines}). This system chooses prunes by doing a brute force search over all pruning schemes on the topological model and selecting the best based on a defined set of 12 tree features. While this method was successful in pruning similar to or better than a human, it is only applicable where brute force search is feasible, such as on cultivars with similar architectures like grapevines. For ``complex" rules and heuristics, implementation failures arise when the quantification metrics used by the autonomous system are different than how the people quantify the same pruning decisions. This disconnect often leads to the autonomous system performing invalid or unusable cuts as validated by experts. Our paper explores ways to quantify pruning decisions that are consistent between people and autonomous pruning systems.

\begin{figure}[h!tbp]
    \centering
    \includegraphics[width=0.95\columnwidth]{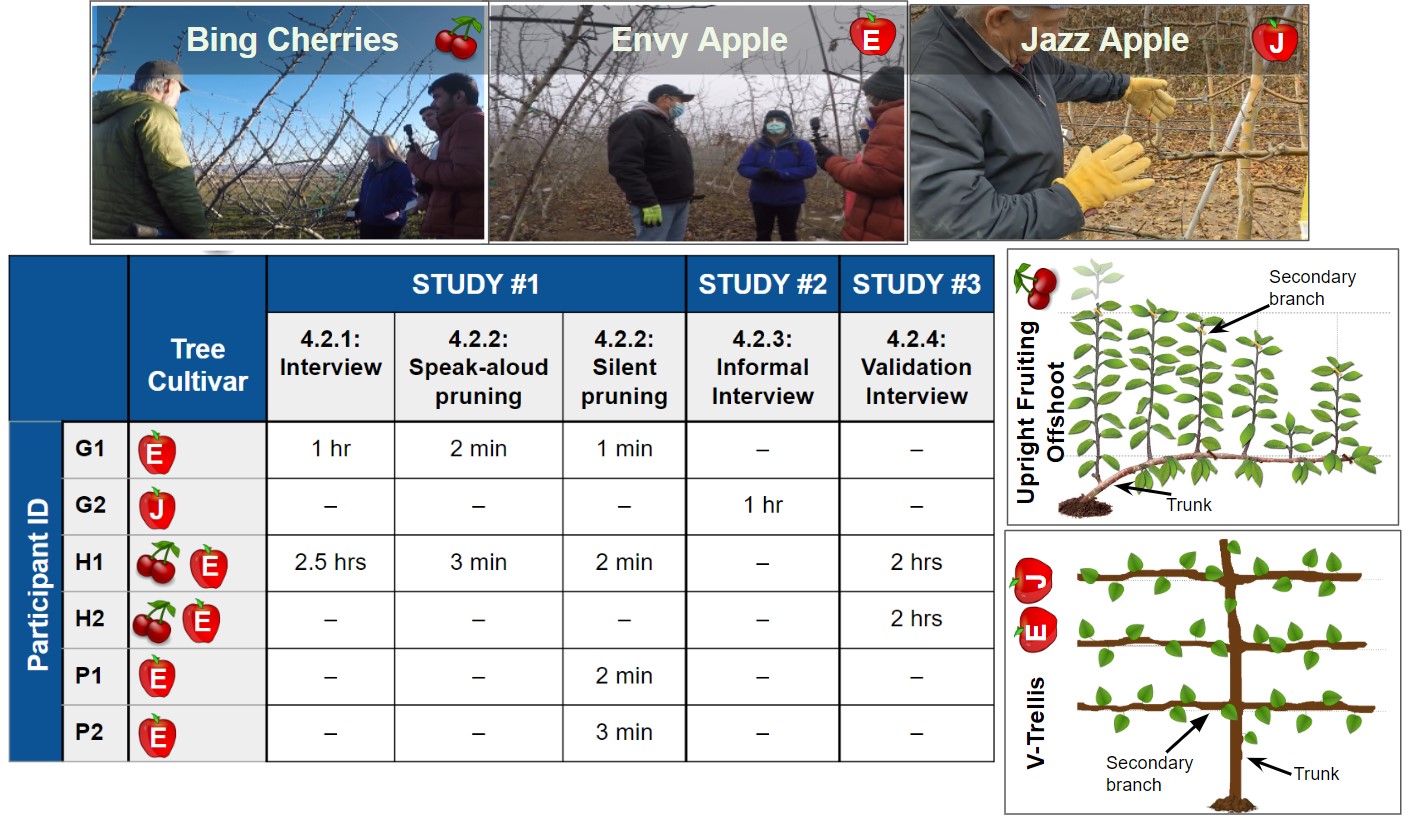} 
    \caption{Breakdown of our three studies based on participant ID and procedure. Participant IDs reflect the person's role, so \textbf{P} is a pruner, \textbf{G} is a grower, and \textbf{H} is a horticulturist. We report how long each participant participated in a specific study procedure and what cultivar and tree architecture (either Upright Fruiting Offshoot or V-Trellis) they interacted with.}
    \label{fig:study_breakdown}
\end{figure}

\section{Pruning Study Methodology: Learning in the Field}
\label{sec:study}


From informal conversations with experts in automation and horticulture, we knew that we would need to interview three types of participants: horticulturists, growers, and pruners. Additionally, we selected two pruning systems {\em a priori}, based on the participants we had access to (UFO cherry and V-trellis apple). When designing the interview protocol we had the following concerns. First, we did not want to make any \textit{a priori} assumptions about what participants deemed ``relevant'' concepts, their terminology, or what went into their decision process before, during, or after pruning. Second, we needed to ground the discussion in the physical tree morphology so we could link terms with actual exemplars. Third --- and somewhat in conflict with the first concern --- we needed to ensure that we captured all relevant information while in the field. Finally, we wanted participants (particularly the pruners) to prune normally  and not ``overthink'' it.

We conducted three studies over the course of one year (see Figure~\ref{fig:study_breakdown}). The first study was an in-field study consisting of a structured interview process largely following the critical decision approach (\citet{klein1989critical}) --- an interview protocol that focuses on identifying decision points and using follow-up questions to elucidate why participants make the decisions they do. We used grounded coding to analyze that study. Our second ``study'' took advantage of a very unique opportunity --- 28 trees manually labeled over 3 days by a grower and his senior pruner. This study included an ad-hoc interview with the grower after the labeling was complete. Our final study was a follow-up study to validate the analysis of the first two studies. 

We first briefly describe the methodology, participants, and protocols used in each study. The first sub-section (Section~\ref{subsec:analysis}) describes what data we collected in our interviews and how we applied grounded coding in general to that data. The following three sub-sections contain details of each study and how grounded coding was applied to those studies.

\noindent {\bf In-field study design (Study 1)}:
This study is an in-field, question-driven protocol with two pruning protocol variations: silent and speak-aloud. We first explained the purpose and style of the study and collected information about the participant. We emphasized the importance of {\em demonstrating}, where possible, the concepts, terms, and cuts on the actual tree. For the speak-aloud protocol, we worked through the questions given in Table~\ref{tab:questions}, prompting follow-up explanations and demonstrations where necessary. For the silent protocol, we simply had the pruners perform cuts as they would normally. This protocol style reflects an ethnographic, fly-on-the-wall observation (e.g.,\citet{chun2020robot,fallatah2020would,silverman2021culture,rothwell2022investigating}). We had originally intended to do a retrospective think-aloud with the pruners but decided not to after witnessing the scale of pruning work (the pruners were working in the field during the interview). 

Key to our interview protocol was developing the list of questions along with a ``checklist'' of additional information/details we wanted to collect for that question. This checklist served as both prompts for follow-up questions and to ensure we did not ``miss'' any data.  


\noindent {\bf In-field (Study 2)}: 
This study is an ad-hoc, in-field study that was made possible through the generous time and effort of one of our growers. The grower, along with one of their most experienced pruners, manually labeled all pruning cuts on 28 V-trellis, Jazz Apple trees (colored twist ties placed around cut points simulating over 1,000 cuts). The grower was informally interviewed by one of the team members, and the conversations were recorded using a cell-phone. Additionally, the trees were fully scanned using both a manual method (cell-phone) and a robotic platform (Azure RGB-D camera mounted on a scanning platform). Since this study took place after the first study was analyzed the interviewer was able to explicitly query the grower about what tree features and high-level goals led to specific cuts.


\noindent {\bf Validation Interview (Study 3)}: 
Following the analysis of the first two studies, we comprised a list of pruning terminology and a set of UFO and V-trellis pruning heuristics. The key to this study was how we presented this discovered pruning terminology and heuristics while encouraging discussion between participants. 

This study is a presentation-based, virtual discussion with two of our team's horticulturists (one being a participant in study 1). For each term, we presented the term alongside the description and, if applicable, a visual followed by the interviewer asking participants if the term was correct and how to modify it. For each heuristic, we presented a mapping of how the branch was growing and why it was cut (i.e., horticultural context)  to specific cut actions using images and video. The interviewer then asked the participants if the heuristic was feasible to use given the context and how to modify the mapping or actions to make the heuristics more usable by people.      


\subsection{Interview and video analysis (studies 1 \& 2)}
Our analysis aims to visually and verbally extract relevant pruning concepts, identify what influences pruning decisions, and discover pruning heuristics by tree cultivar. Furthermore, we wanted to avoid making \textit{a priori} assumptions and instead have the data to guide the analysis process. 

We created an analysis technique (based on grounded coding) to extract pruning decisions. Our analysis uses more than six hours of video data collected from interviews and pruning observations with all audio converted to transcripts. For all analyses, two researchers worked concurrently in a shared excel sheet to extract and code transcript text and sections of video, and whenever possible, link images that visually describe the transcript quotes. Researchers resolved any disputes throughout this process by discussing why the data satisfied the extraction criteria based on the type of analysis being performed.   

Our analysis is divided into two sections: pruning terminology (Section \ref{sec:terminology}) and heuristics (Section \ref{sec:heuristics}). For the pruning terminology, two researchers extracted participant transcript quotes and video instances from all interviews and pruning observations when i) the participant physically identified a tree feature by gesturing (with hands or loppers); ii) the participant demonstrated an action and provided a corresponding name; or iii) the participant gave a definition to a corresponding term like ``biennial bearing is where it is very productive one year, and then those fruit(s) inhibit flowers for the next year.'' All terminology the researcher's identified was saved with an associated term (e.g., ``bud'') and the definition or term and video instance (when possible) in the shared excel sheet. 

For the pruning heuristics analysis, the same researchers extracted video clips where the participant either made a physical pruning cut or the tree had a colored tag. Key to this analysis was, for each branch physically pruned or tagged branch, we added generalized and cultivar-specific descriptions about the branch (e.g., \textbf{Branch Vector}) in different columns representing different tree characteristics (e.g., length) in a shared excel sheet. The labels and corresponding descriptions were agreed upon by both researchers either before or during the analysis. We also extracted sections of transcripts when participants provided explanations about prunes or what influenced their pruning decision. Finally, using the extracted videos and branch descriptions, one researcher independently mapped the total number of cuts to various groupings of generalized and cultivar-specific labels; the same labels used to describe the branch when we extracted video clips. These mappings represent pruning heuristics discovered from real-world cuts to the three cultivars, or more specifically, the two tree architectures.    

The remainder of this section is as follows: Subsection~\ref{sec:participant} details our selection process of participants and cultivars. Subsection~\ref{sec:data_collect} details our interview protocols and data collection methods for all three studies; the formative study interview in Subsection~\ref{subsubsec:interview} and pruning protocols in Subsection~\ref{subsubsec:protocol}; the informal grower interview procedure in Subsection~\ref{subsec:tagged}; and the validation interview in Section~\ref{subsec:valid}. Finally, Subsection~\ref{subsec:analysis} defines our grounded coding analysis technique to identify pruning terminology and heuristics.

\subsection{Selection of Participants and Cultivars}
\label{sec:participant}
Across all our studies, we received consent and interacted with six people, two from each pruning stakeholder category (see Figure~\ref{fig:study_breakdown} ``Participant ID''). We recruited participants through established contacts of our research team's horticulturists. The horticulturists are members of our research team who specialize in efficient orchard architectures and agricultural fruit mechanization. The two growers are contacts of the horticulturists and research group (WTFGA) who are interested in automation and modern techniques in agriculture. Finally, the grower in the in-field formative study (participant $ID = G1$) selected the two pruners on the day of our visit based on pruning skill level. 

We selected the cultivars based on guidance from our horticulturalist experts. The Bing Cherry cultivar is from an affiliated university institute's research site and utilizes an Upright Fruiting Offshoot (UFO) architecture, co-developed by the interviewed horticulturist (\citet{whiting2011ufo, whiting2018precision}). The UFO architecture was designed to be both easy to prune and to harvest (\citet{whiting2018precision,du2011mechanical,whiting2011ufo,lang2015cherry}). The Envy and Jazz Apple cultivars both employ the V-trellis architecture, originally designed to simplify harvesting, with fruit easily reachable from a moving vehicle (\citet{robinson2006tall}). This architecture trades-off pruning complexity for yield-control and harvesting speed. Although not originally designed for automation, both architectures are essentially ``planar'', making it feasible to develop robotic pruning systems for them. 

\subsection{Interview Protocol and Data Collection}
\label{sec:data_collect}

This subsection defines our interview procedures, pruning protocols, and data collection methods for our in-field formative study, the informal grower interview with tagged trees, and the pruning decision validation interview. For all participant interactions, we captured video and audio recordings and later transcribed them for grounded coding analysis. 


\subsubsection{Study 1: In-the-Field Interviews}
\label{subsubsec:interview}

The participant interview followed a semi-structured format and lasted approximately one hour, with the first 45 minutes focused on pruning and the remaining 15 minutes on robotic expectations and workflow. Initial questions are given in Table~\ref{tab:questions}. The questions address decision-making strategies, pruning considerations and instructions, evaluation techniques, and preferred robot interactions. Our team also asked follow-up questions to clarify concepts or terminology throughout the interview and encouraged participants and study members to point at the tree. Examples of follow-up question structures are:
\begin{enumerate}
    \item What do you mean when you say [terminology]?
    \item Can you point to [terminology] on the tree?
    \item Can you demonstrate [pruning action]?
    \item Why did you make [pruning decision]?
    \item What influenced you to make/take [pruning decision/action]?
\end{enumerate}

Our in-the-field research team consisted of five people (two interviewers, three recorders), with two cameras and two microphones. The primary interviewer asked the prompt and follow-up questions. The secondary interviewer made note of which items on the check-list had been covered and communicated with the primary interviewer if any check list items were missing. We used a dedicated audio recording device (recorder 1) plus one Go-Pro to record the entire scene from a tripod (recorder 2), while the second Go-Pro was manually moved around to ``zoom in'' on any features the grower mentioned/the pruners cut (recorder 3). 

Participants and cultivars: We interviewed four participants —-- one horticulturist, one grower, and two pruners —-- in Prosser, Washington, in January 2022. The cultivars were Envy V-trellis apples and Bing cherries in an Upright Fruiting Offshoot (UFO) architecture. Total interview time was approximately four hours.

\begin{table}[h!tbp]
\begin{tabular}{ |p{13.5cm}| }

\hline
\rowcolor{applegreen}
\textbf{Relevant Pruning Concepts} \\
\hline
- What features do you use to determine age of a branch? \\
- What is considered a healthy branch? What features were you looking at/considering? \\
- How do you determine a vegetative versus flowering bud? \\

\hline
\rowcolor{applegreen}
\textbf{Pruning Heuristics and Instructions} \\
\hline
    - What instructions do you give/would you give to the pruners?
    - Do the instructions vary with tree age? \\
    - Do you have specific rules for pruning? \\
    - How specific do you need to be in giving instructions? \\
    - Do you ever watch someone prune and then provide additional instructions? \\
    - How much variation do you see in pruners? What does it look like? \\
    - What specifics about this block are you using to make your instructions? Root stock? Prune type? Previous yields?\\
    - Does the age of the branch matter for pruning? \\
\hline
\rowcolor{applegreen}
\textbf{Pruning Decisions and Strategies}  \\
\hline
     - What do you consider when you choose a pruning strategy. What are you thinking/making decisions about? \\
     - What is your process for deciding how to prune this block? How much of this varies per block? \\
     - What are the goals of pruning? What are you trying to achieve by pruning? \\
     - Do you have a wider strategy for pruning? \\
     - Are there different types of pruning you might do? \\
\hline
\rowcolor{applegreen}
\textbf{Pruning Evaluation} \\
\hline
    - How do you evaluate the resulting prune? \\
    - How do you evaluate a good versus bad prune? \\
    - How much evaluation of a good or bad prune can yon figure out from just looking at the tree? \\
    - How much of evaluating a good or bad prune is determined by evaluating later? \\

\hline
\rowcolor{applegreen}
 \textbf{Robot Expectations and Preferred Interactions}  \\
 \hline
    - How do you envision interacting with a robotic pruning system? \\
    - How would your pruners interact with a robotic system? \\
    - What do you think the robot would do [for pruning]? \\
    - How would you like to see prunes? Before and after trees (digital)? Showing cuts before they're made? Let the pruners spray paint marks (robot system cuts)? \\
    - How comfortable are you with letting the robot make pruning decisions? \\
    - What would make you more comfortable with the robot? Seeing before and after trees images (digital)? Seeing it do the cuts? \\
\hline

\end{tabular}
 \caption{List of questions utilized in the In-the-Field formative study participant interviews described in Section \ref{subsubsec:interview}}
\label{tab:questions}
\end{table}

\subsubsection{Silent and Speak-aloud Pruning Protocols}
\label{subsubsec:protocol}

In order to observe pruning in the field, we developed two pruning protocols: silent and speak-aloud. For the \textit{speak-aloud} pruning protocol, the participant 
demonstrated pruning while providing explanations, such as  why they were making that particular cut and their cut strategy. During this protocol, our primary and secondary interviewer interjected questions to ensure all unknown explanations and visual terms were grounded in actual tree morphology. In contrast, for the \textit{silent} pruning protocol, the participant pruned without providing explanations and no interruption from our research team.   



Participants and cultivars across both pruning protocols are the same as the in-the-field interviews described above. The grower and horticulturist utilized the speak-aloud protocol, while the two pruners used the silent pruning protocol. We recorded a total of 13 minutes of video across both protocols (silent: 8 minutes; speak-aloud: 5 minutes).

\begin{figure}
    \centering
    \includegraphics[width=0.85\linewidth]{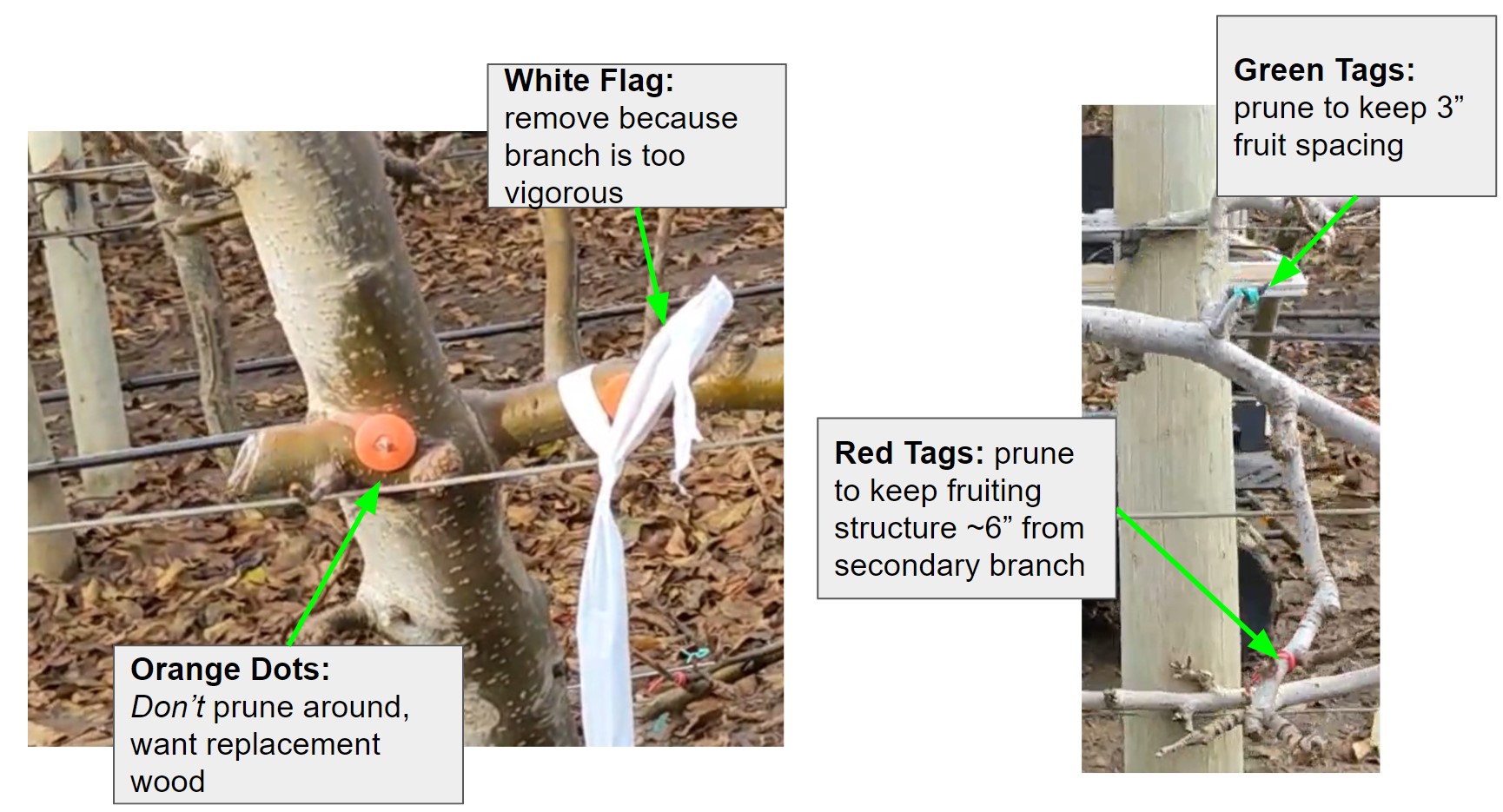}
    \caption{Example of a tagged tree with the four colored pruning tags. Each colored tag corresponds to a different pruning action. Green: Fruit spacing, Red:  Self-shading or out-of-canopy growth.}
    \label{fig:tagged_tree}
\end{figure}

\subsubsection{Study 2: Informal grower interview with marked trees}
\label{subsec:tagged}

Research members conducted an additional pruning decision interview in January 2023 with a grower and 28 tagged apple trees (see Figure~\ref{fig:tagged_tree}). Tag placement was a collaborative effort between the grower and an expert pruner. Different pruning decisions were marked with red or green tags, white tags, white flags, and orange dots. Red tags indicated pruning decisions to maintain the fruiting structure, or limiting outward growth from the secondary branch between six and eight inches. Green tags marked prunes for maintaining two inch fruit spacing between buds. Likewise, white flags identified branches that were too vigorous that need to be removed. White tags were branches that were too close to the support branch and would produce shaded fruit or fruit that would rub against the branch. Finally, orange dots indicated places within a six-inch radius from the tag a pruner should avoid pruning so the tree can grow replacement wood. This tagging represented highly detailed and precise pruning decisions, far more than hired pruners would utilize ``in practice'' given time constraints. Questions asked by the research team were largely around clarifying tag placements and why that pruning cut corresponded to a specific color. Additionally, the grower answered a series of questions about the overall architecture (angle of trellis, spacing) and the 3-5 year process of establishing the mature orchard. The questions are similar to those asked in study 1 (see Table~\ref{tab:questions}).

\subsubsection{Study 3: Pruning Decision Validation Interview}
\label{subsec:valid}

One research member conducted a follow-up interview over Zoom with two horticulturists (one being the horticulturist initially interviewed in study 1) to validate the extracted pruning terminology and heuristic for Upright Fruiting Offshoot and V-trellis and further connect local pruning heuristics to high-level horticultural concepts. The researcher, acting as the interviewer and discussion moderator, oriented the horticulturists to the discovered pruning decision structure in the first five minutes, followed by twenty minutes presenting terminology and definitions alongside visual examples for Cherry and Apple trees. For most of the interview time (i.e., 70 minutes), the interviewer presented individual pruning heuristics captured from analysis of the first two studies (see Section~\ref{subsec:heuristic_analysis}) with accompanying diagrams and real-world pruning examples (45 minutes for Apples, 25 for Cherries). During the remaining time, the interviewer asked the horticulturists to discuss pruning contexts that can influence pruning decisions not currently addressed or considered and workflow and robotic considerations.  

Table~\ref{tab:valid_questions} reports the questions asked by the interviewer based on what information was being validated for both pruning terminology and heuristics. The interviewer formulated each question to encourage discussion amongst the interviewer and horticulturists about the validity of the pruning heuristics and recommendations for improving the pruning decision by modifying labels and terminology. We report our final pruning terminology set and discovered pruning heuristics, incorporating any suggestions recommended during this interview, in the following sections. 


%
%

\begin{table}[!h]
\begin{tabular}{ |p{13.5cm}| }

\hline
\rowcolor{applegreen}
\textbf{Pruning Terminology} \\
\hline
- Would you change the term definition in any way? If yes, how? \\
- Is the definition used for the term accurate? \\
- What other terms do you feel are missing from the list shown? \\

\hline
\rowcolor{applegreen}
\textbf{Pruning Decisions and Horticultural Goal to Heuristics Mapping} \\
\hline
    - Is the current horticultural concept this heuristic mapped to correct/accurate? Is there another concept you would add? \\
    - Does this mapping form \textit{Pruning Branch Labels} to \textit{Physical Cut} labels easy to follow? \\
    - Would you modify this heuristic in any way? How? Why?  \\
    - Is there a pruning context you feel is missing that would modify this heuristic? \\
\hline
\rowcolor{applegreen}
\textbf{Pruning Contexts} \\
\hline
    - What are other ways in which planters vary pruning to the particulars of their fields? How much does that vary across particular people, crops, etc? \\
    - Are there any ways we can parameterize the rules more, i.e., allow for more input by people? \\
    - Are there opportunities for parameterization, interfaces, algorithms to assist in specifying these?\\
\hline
\rowcolor{applegreen}
\textbf{Workflow} \\
\hline
    - Do you have any insights on how you might expect farmers to integrate mechanical pruning and/or pruning interface customization/intelligence going into the future? \\
    - In terms of information flow, what might be the role of managers… humans in the field… co-labors? \\
    - Any other thoughts on interfaces or algorithms? \\
    - Anything else you wanted to say or that you hoped we would ask? \\
    - We think it would be really important to show growers (on a screen with virtual trees or by painting branches) actual pruning cuts the system would propose, rather than “rules” per-se. Does this make sense? \\
    - What “knobs” do you think growers would like to control with a pruning interface? Number of buds? More or less vegetative growth? \\
\hline

\end{tabular}
 \caption{List of questions (grouped by discussion topic) utilized in the validation interview described in \ref{subsec:valid}}
\label{tab:valid_questions}
\end{table}

\subsection{Grounded Coding Analysis Approach}
\label{subsec:analysis}


In this subsection we define grounded coding, also called grounded theory (\citet{bryant2007sage, pandit1996creation,charmaz2014constructing}). In the following subsections we apply this approach to extract the pruning terminology (Subsection~\ref{subsec:terminology_analysis}) and heuristics (Subsection~\ref{subsec:heuristic_analysis}). 

Grounded coding is a multi-step analysis technique widely used in various social sciences, such as anthropology and psychology (\citet{von2021two,henwood2003grounded,mohajan2022development}), to analyze semi-formal and informal interviews of subject experts. This analysis technique and how we applied it consists of three steps: 

\begin{enumerate}
    \item \textbf{Extracting data:} in this step we identified transcript quotes and video clips that share a common trait relating to one or more of our research questions and follow-up prompts.
    \item \textbf{Open coding:} in this step we add descriptive labels to the extracted data that define the perceived connection between the data and research question(s).
    \item \textbf{Axial coding:} in this step we group subsets of labels created in step two to capture holistic relationships for each research question. 
\end{enumerate}

Each of the three steps, and their connection to the targeted research questions, is how the analysis stays ``grounded'' without having \textit{a priori} knowledge of the data. Instead, the open and axial coding process reveals relevant concepts that researchers may not have identified or considered using different analysis techniques. 


For all analyses, two study members worked concurrently in a shared excel sheet. The sheet had a dedicated column for referencing which transcript or video the researcher extracted the text or visual data from and a column to store the extracted data. We added additional columns in the data sheet to help describe the extracted data, but these columns are dependent on the analysis. For pruning terminology, one column was added for textual definitions of the extracted terminology. For pruning heuristics, these columns came from pruning terminology reported in section~\ref{sec:terminology} that represent visual tree contexts (e.g., tree geometry) and cut actions.  Both researchers resolved disputes in what and how to extract and label data by discussing either how the data fit the extraction criteria (for pruning terminology) or why the data matched the associated terminology description (for heuristics). We report the exact implementation of grounded coding analysis for pruning terminology and heuristics in the following subsections. 

As a reminder, the extracted terminology and heuristics were further validated/refined in our third follow-up study.

\subsubsection{Pruning Terminology Analysis}
\label{subsec:terminology_analysis}

Due to the complexity of the data (visual and audio) and the need to link definitions and descriptions to visuals, we used three passes for the {\em extracting data} step of the grounded coding (visual, visual plus terminology, and definitions). In all cases, two study members stored the resulting extracted data as text and images in a shared excel sheet containing three columns.

\begin{enumerate}
    \item \textbf{Term}: a keyword or phrase used to describe an action or object. 
    \item \textbf{Definition}: a verbal description of the action's or object's appearance (e.g., general motion or shape) and purpose.
    \item \textbf{Visual Cue}: example image/video segment demonstrating the term.  
\end{enumerate}


The first pass method consisted of identifying and extracting sections of transcripts and video where the interviewee physically pointed or gestured to a visual tree feature with their hand or loppers while providing a description. For example,  when the horticulturalist identified a bud by pointing to it and saying, ``buds, they're covered in frost here, but this is a bud.'' 

\begin{figure}[!h]
    \centering
    \includegraphics{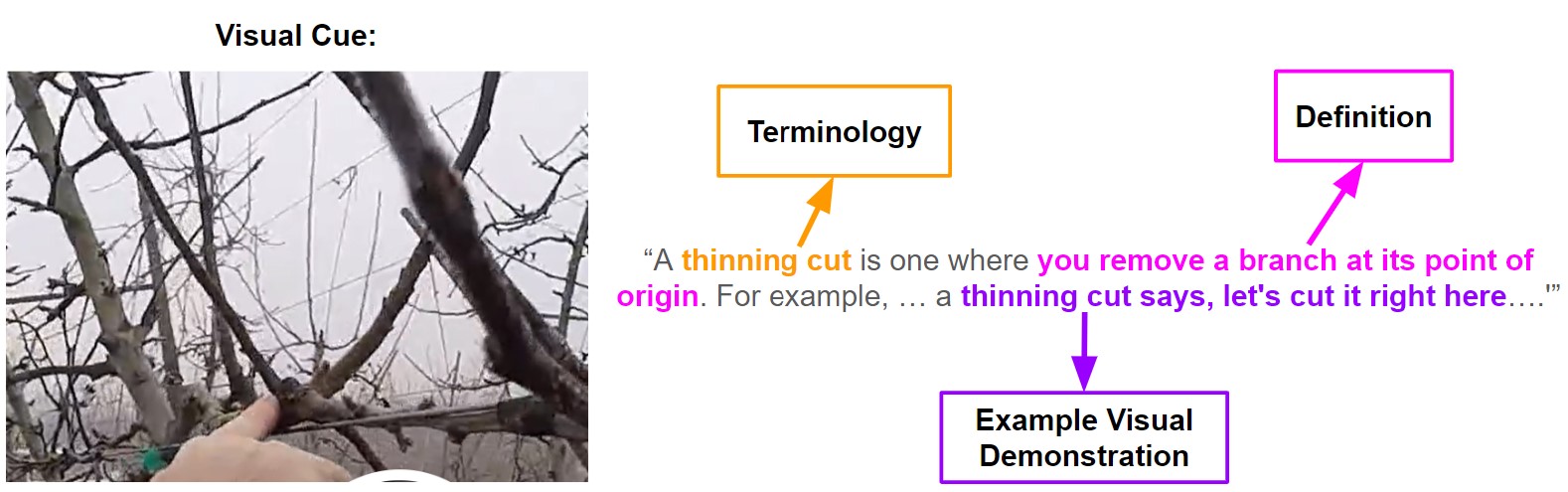}
    \caption{An example of how terminology was extracted from transcript quotes and visual cues during the in-field interviews. The discovered term --- ``thinning cut'' --- is represented in orange with its corresponding definition in pink. Finally, an example visual representation of the cut explained and demonstrated by the participant by pointing to the branch is shown in the left side of the image labeled ``Visual Cue'' and the accompanying text in purple.}
    \label{fig:example_term}
\end{figure}

The second pass extracted video and transcript quotes whenever a participant demonstrated an action and provided a corresponding name. As an example, the horticulturist (while describing a cut) would point or use their hands as ``scissors'' to demonstrate where the lopper would be placed for a ``thinning cut'', as shown in Figure~\ref{fig:example_term}.

Finally, the third pass looked for quote instances when a participant provided an accompanying definition or example descriptor to a corresponding term. For example, ``biennial bearing is where it is very productive one year, and then those fruit(s) inhibit flowers for the next year''. This quote identified the term ``biennial bearing'' and provided a definition for that term. 

After extracting the data, the next step in the grounded coding process is {\em open coding}. In this step, we created definitions using explanations and video given by participants during interviews and pruning observations for each term. When participants provided multiple explanations for a given term, both researchers worked together to create a definition that captures common themes across all provided explanations. For example, for the term ``thinning cut'' we defined the corresponding definition as ``a type of cut performed by a person which removes a branch at its base." The definitions, therefore, represented our label set.      


The final step is the {\em axial coding}. Here, we grouped each term and its corresponding definition into sub-categories based on similar characteristics such as tree features or physical cuts. For example, all terminology that described the direction a branch was growing were all grouped together as a \textbf{Branch Vector}. Lastly, two study members grouped discovered sub-categories into hierarchical levels based on contextual and visual inputs a robotic pruning system would need and physical outputs or actions the robots might perform. We provide the final set of terminology and the hierarchical levels in Section~\ref{sec:terminology}, and in Tables~\ref{tab:tree_context}--\ref{tab:physical_cut}. 
\subsubsection{Pruning Heuristics Analysis}
\label{subsec:heuristic_analysis}

We analyzed participants' pruning processes in two phases: i) extracting and labeling physical pruning cuts from video of the two pruning protocols, informal interviews with marked trees, and in-the-field interviews, and ii) constructing heuristics for each cultivar based on how participants explained their pruning decisions.   

For extracting and labelling pruning data, two researchers recorded all data as text in a shared excel sheet. This excel sheet contained 17 qualitative codes (one code per column) that recorded the data's origin (i.e., which video and the timestamp) and characteristics of the branch, such as the number of buds, \textbf{Branch Length}, or \textbf{Branch Vector}. All columns --- and for which tree cultivars they were applicable to --- were agreed upon by both researchers before or during the analysis process. Furthermore, for certain qualitative codes, researchers also agreed on the associated visual cues. For example, for the \textbf{Branch Vector} column, researchers added one of six agreed upon descriptors --- ``Parallel to the Wire," ``Into the Row'', ``Into the Interior'', ``Out of the Row'', ``Up from Wire'', and ``Down from Wire'' --- based on the tree architecture. We report the set of columns and their associated descriptions in Section~\ref{sec:terminology}.   

\subsubsection{Extracting and Labelling Pruning Data} 
Our {\em extracting data} step consisted of finding video segments where there were instances of either a physical pruning cut --- the participant's loppers came in contact with a branch and that branch was cut --- or tagged branches --- branches visibly marked with at least one colored tag defined in Section~\ref{subsec:tagged}. When either researcher identified a physical prune or tagged branch, it was documented in the excel sheet with which video (video name) and the time stamp of when  the prune or tag happened. All instances identified by both researchers represent individual rows in the shared excel sheet.

\begin{figure}[h!tbp]
    \centering
    \includegraphics[width=0.97\columnwidth]{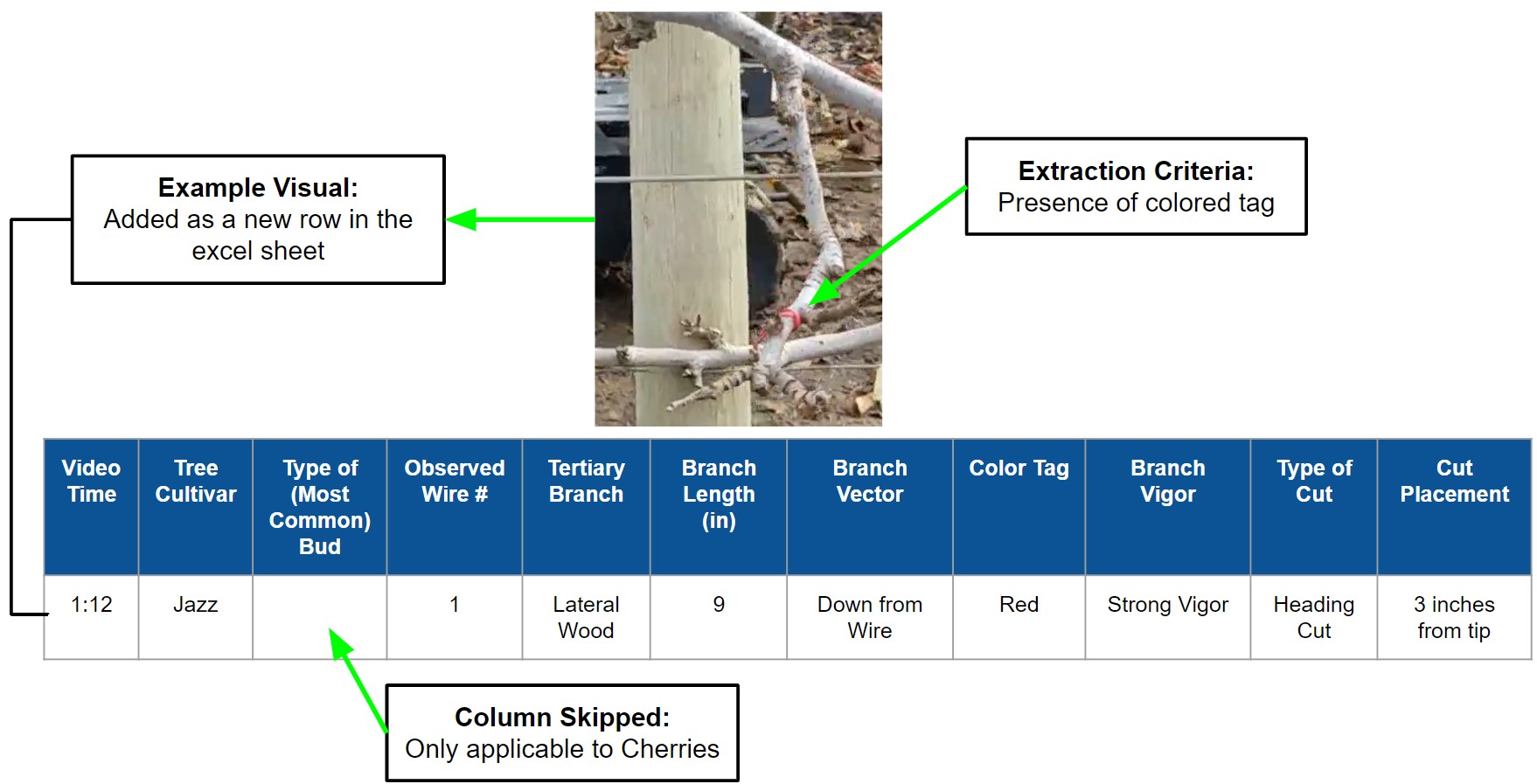}
    \caption{An example of our {\em extracting data} and {\em open coding} step for pruning heuristic analysis. We extracted a branch from a Jazz Apple video segment due to the presence of a colored tag (red). The extracted branch was added as a new in the excel sheet and visual cues filled in by one researcher in the appropriate columns for that cultivar's architecture (Apples: V-Trellis; Cherries: UFO).}
    \label{fig:prune_analysis}
\end{figure}

For the {\em open coding} step, one researcher filled in visual cues in the associated qualitative code (column) for each extracted data instance (row). We provide an example of open coding labelling with a subset of columns in Figure~\ref{fig:prune_analysis}. The qualitative code was left blank in the excel sheet if it was not relevant for that cultivar's architecture. For example, we only used \textbf{Observed Wire Number} when describing cultivars with the V-trellis architecture (Apples) and not for Upright Fruiting Offshoot (Cherries).

Finally, our {\em axial coding} step involved moving excel columns left or right based on label groupings and order of relevance. We report the complete set of qualitative labels (terminology and visual cues) for each cultivar in Section~\ref{sec:terminology} (see Tables~\ref{tab:tree_context}--\ref{tab:physical_cut}.).

\subsubsection{Constructing Heuristics}

\begin{figure}
    \centering
    \includegraphics[width=0.97\linewidth]{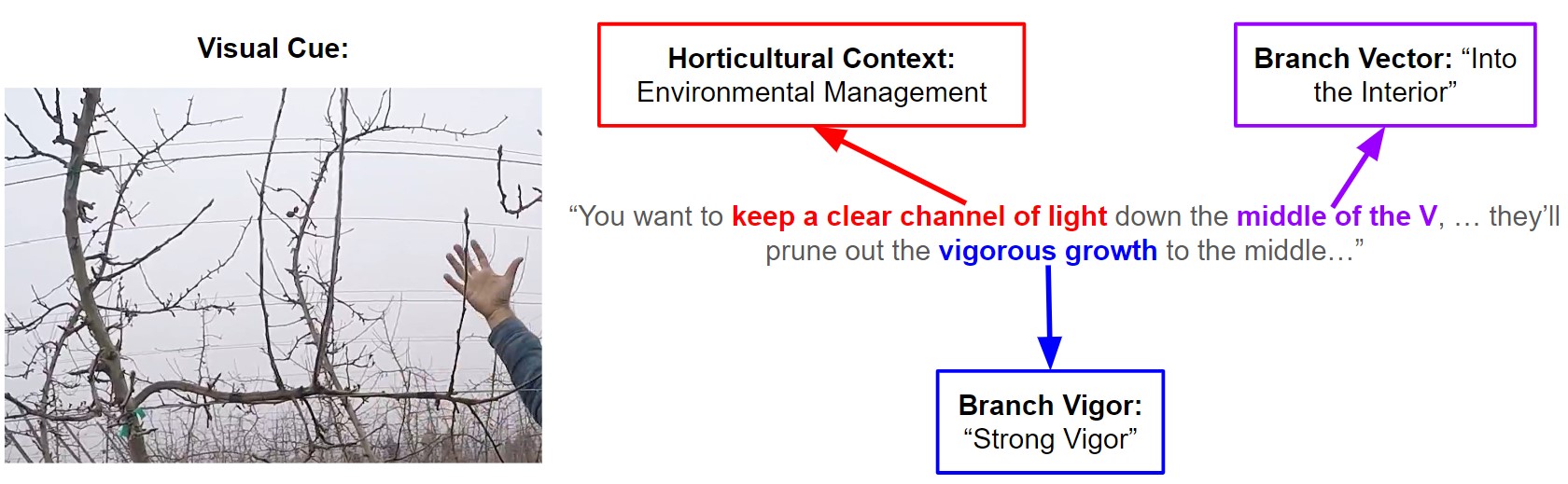}
    \caption{We utilized participant quotes to capture pruning heuristics syntax by identifying terminology (excel columns) in the text. }
    \label{fig:analysis_text}
\end{figure}

To construct heuristics for each tree cultivar, two study members analyzed participant explanations provided during pruning observations and in-the-field interviews to discover connections between qualitative codes. Specifically, we rewrote explanations with the corresponding qualitative code or terminology sub-categories. Figure~\ref{fig:analysis_text} shows an example mapping of a participant quote to terminology categories and the corresponding visual cue as shown in purple and blue. The pruning explanation mappings provided a starting point for performing an analysis to map input labels to output labels. It also helped identify the global context or horticultural goal of why the participant made a cut (shown in red) and which labels were more or less critical, unnecessary, or redundant for each tree cultivar and pruning decision.

Utilizing these discovered mappings, one study member numerically analyzed pruning cuts by mapping the percentage of occurrences for specific input label descriptors to output label descriptors in ranked order. The study member drew lines connecting specific input-output label descriptors if a) every instance of the input-output mapping (i.e., numerical analysis returning 100\%) maps to this decision or b) the input-output mapping returns the maximum percentage after utilizing all labels. Section~\ref{sec:heuristics} reports the example heuristics, ordered by cultivar. 

\section{Results}
\label{sec:results}

We present the results of our grounded coding analyses for pruning terminology (Subsection~\ref{sec:terminology}) and heuristics (Subsection~\ref{sec:heuristics}) in the following two subsections. 

For pruning terminology, we report the four hierarchical terminology categories discovered from our analysis and the 16 associated terms, definitions, and visual representations extracted from six hours of interview and pruning observation video and audio transcripts. These terms and definitions are used in formulating our heuristics by grouping visual-based input tree features to action-based outputs. 

For pruning heuristics, we report the three most common contexts that influence pruning decisions. We also present a set of seven pruning heuristics that map from these contexts to specific cut decisions categorized from 661 distinct physical pruning cuts and tagged tree instances (Bing Cherries, N=124; Envy Apples, N=229; Jazz Apples, N=308) across five participants --- two growers, two pruners, and one horticulturist --- over 13 minutes of pruning observation video (Bing=5 min, Envy=6 min, Jazz=2 min). Finally, we discuss four pruning visual-spatial tasks participants use when pruning and how they affect cut decisions.  

%
%
\subsection{Results: Pruning Terminology and Definitions}
\label{sec:terminology}
 
In this section we present the terminology and definitions that were extracted from the raw interview footage (Section~\ref{subsec:terminology_analysis}) and refined in the Validation study (Section~\ref{subsec:valid}). In the {\em axial} coding step these terms were organized into a hierarchy four levels deep with four top-level categories. We present each of these four categories in turn, along with their sub-category terms.

%
%
\begin{table}[h!tbp]
    \small
    \centering
    \begin{tabular}{|p{3cm}|p{7cm}|p{2.5cm}|}
    \hline
        \rowcolor{CherryBlossomPink}
        \textbf{Label} & \textbf{Defintion} & \textbf{Example} \\
       \hline
        \textbf{Tree Cultivar} & The biological cultivar of the tree & Envy Apple \\
        \hline
       \textbf{Tree Architecture} & Verbal description of the current plant architecture of the tree & Upright Fruiting Offshoot (UFO) \\ 
        \hline
        \textbf{Root Stock}& The biological classification of the tree's root structure & Algae-41 \\
        \hline
    \end{tabular}
    \caption{Tree Context has three corresponding terminology — tree cultivar, tree architecture, and root stock — that represent high-level horticultural concepts and knowledge of the tree cultivar being pruned. We provide definitions and example classifications for each term.}
    \label{tab:tree_context}
\end{table}

\subsubsection{Tree Contexts}
This category contains general terms for high-level horticultural concepts and information about the tree cultivar. These terms do not directly influence individual pruning decisions. However, they do specify how specific tree cultivar characteristics influence high-level pruning goals. Table~\ref{tab:tree_context} shows the three terms at the next level of the hierarchy along with their definitions

\textbf{Tree Cultivar} refers to the biological classification of the cultivar. Our study included three different tree cultivars: Bing Cherries, Envy Apples, and Jazz Apples.

\textbf{Tree Architecture} refers to the architecture of the tree. Our study included two tree architectures: Upright Fruiting Offshoots (UFO) and V-trellis (apples). Figure~\ref{fig:study_breakdown} shows an abstract representation of these two architectures. 

\textbf{Rootstock} refers to the biological classification of the tree's root stock (the cultivar below the graft union). In our studies, the apple trees used  Algae-41 for the apple root stock, which is a dwarf root stock. 

\begin{table}[h!tbp]
    \small
    \centering
    \begin{tabular}{|p{2cm}|p{2cm}|p{7cm}|p{2cm}|}
    
    \hline
    \rowcolor{CherryBlossomPink}
    \textbf{Label} & \textbf{Cultivar} & \textbf{Definition} & \textbf{Example Label} \\
    \hline

    \textbf{Tertiary Branch Type} & General & Biological classification of the tertiary branch, e.g., whether it will grow fruit in cherries & Lateral Spur Wood, Lateral Vegetative Wood \\
    \hline

    \textbf{Branch Length} & General & An estimation of the length of the branch given in inches before pruning & e.g, 8, 10, 14\\
    \hline

    \textbf{Branch Vector} & General & The \textit{growth direction of the branch} relative to the main branch or overall plant structure & e.g., Into the interior, Parallel to wire\\
    \hline

    \textbf{Branch Vigor} & General & A description of the \textit{vigor} of the current branch & Weak, Strong \\
    \hline
    \hline

    \textbf{Type of (Most Common) Bud} & Cherry & A description of the \textit{most prevalent bud} on the branch being pruned & Fruiting Spur, Vegetative \\
    \hline

    \textbf{Number Fruiting Spurs before/after cut} & Cherry & The total number of fruiting spur buds present on the branch before/after pruning & e.g., 2, 4, 5 \\
    \hline

    \textbf{Number Buds before/after cut} & Apple & The total number of buds present on the branch before/after pruning & e.g., 0, 1, 2 \\
    \hline

    \textbf{Observed Wire Number} & Apple & Numerical representation of the wire number closest to the branch & e.g., 1, 2, 3 \\
    \hline

    \end{tabular}
    \caption{There are seven pruning branch labels that represent different tree features that influence pruning decisions. These labels can be divided into two categories: generalized and cultivar-specific. Generalized labels are labels that apply to all three tree cultivars (i.e., Bing Cherry, Envy Apple, and Jazz Apple) and is represented by ``General" in the \textbf{Cultivar} column, while the cultivar-specific only apply to one or two.}
    \label{tab:pruning_branch}
\end{table}

%
%

\subsubsection{Pruning Branch Terms}
This category contains eight terms that define either geometric properties of the branch(es) or biological ones. These properties are the ones that are used to identify which branches to prune, and where to prune them. Four of these terms were used in all cultivars, while the other four were used when discussing a specific cultivar (apple or cherry). This reflects the tree biology and architecture.~\footnote{In cherries fruiting versus vegetative buds are distinct, while in apples a bud may contain a mix of leaves and flowers.} The terms are summarized in Table~\ref{tab:pruning_branch}.

\begin{figure}[h!tbp]
    \centering
    \includegraphics[width=0.85\columnwidth]{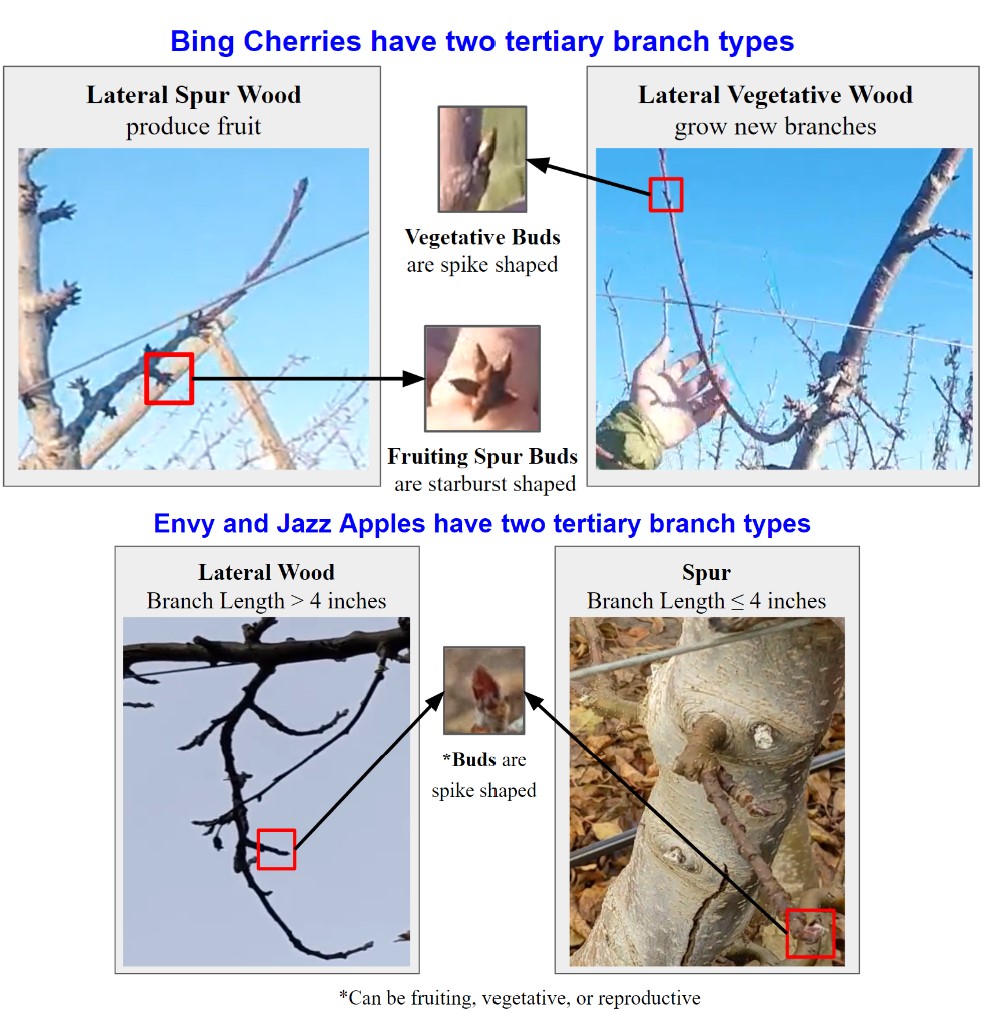}
    \caption{Bing Cherries possess two tertiary branch types: Lateral Spur Wood and Lateral Vegetative Wood. ``Lateral Spur Wood'' are branches that will produce cherries, while ``Lateral Vegetative Wood'' will grow new branches. These branch types are distinguishable by the shape of the end bud. Fruiting Spur Buds are shaped like starbursts while vegetative buds are singular spikes. Envy and Jazz Apples also possess two branch types --- ``Lateral Wood" and ``Spurs" --- distinguishable by branch length (Spurs $\leq$ 4 inches, Lateral Wood $>$4 inches). Both branch types possess spike-shaped buds which can be either fruiting, vegetative, or reproductive.}
    \label{fig:branch_type}
\end{figure}

\textbf{Tertiary Branch Type [General]} refers to the biological classification of the tertiary branch. In Bing Cherries, there were two tertiary branch Types: ``Lateral Spur Wood'' and ``Lateral Vegetative Wood''. ``Lateral Spur Wood" refers to starburst-shaped buds which will produce cherries, while ``Lateral Vegetative Wood'' produces new branches (see Figure~\ref{fig:branch_type}, top row). 

Envy and Jazz Apples also possessed two tertiary branch types: ``Lateral Wood'' and ``Spur'' (see Figure~\ref{fig:branch_type}, bottom row). These branch types differ in length. ``Spur" tertiary branches are typically short (less than 4 inches), while ``Lateral Wood" is longer with fewer buds.     

\textbf{Branch Length [General]} refers to the total length of a branch from its base to the tip. The value associated with this term is typically a numerical approximation since  directly measuring each branch is not feasible in practice. Length of a branch is one feature used to identify which branch to remove.

\textbf{Branch Vector [General]} refers to the growth direction of a branch relative to the main branch or architecture/overall plant structure. The value associated with this term is typically qualitative. This angle is another feature used to define which branch to remove. 

\begin{figure}
    \centering
    \includegraphics[width=0.95\linewidth]{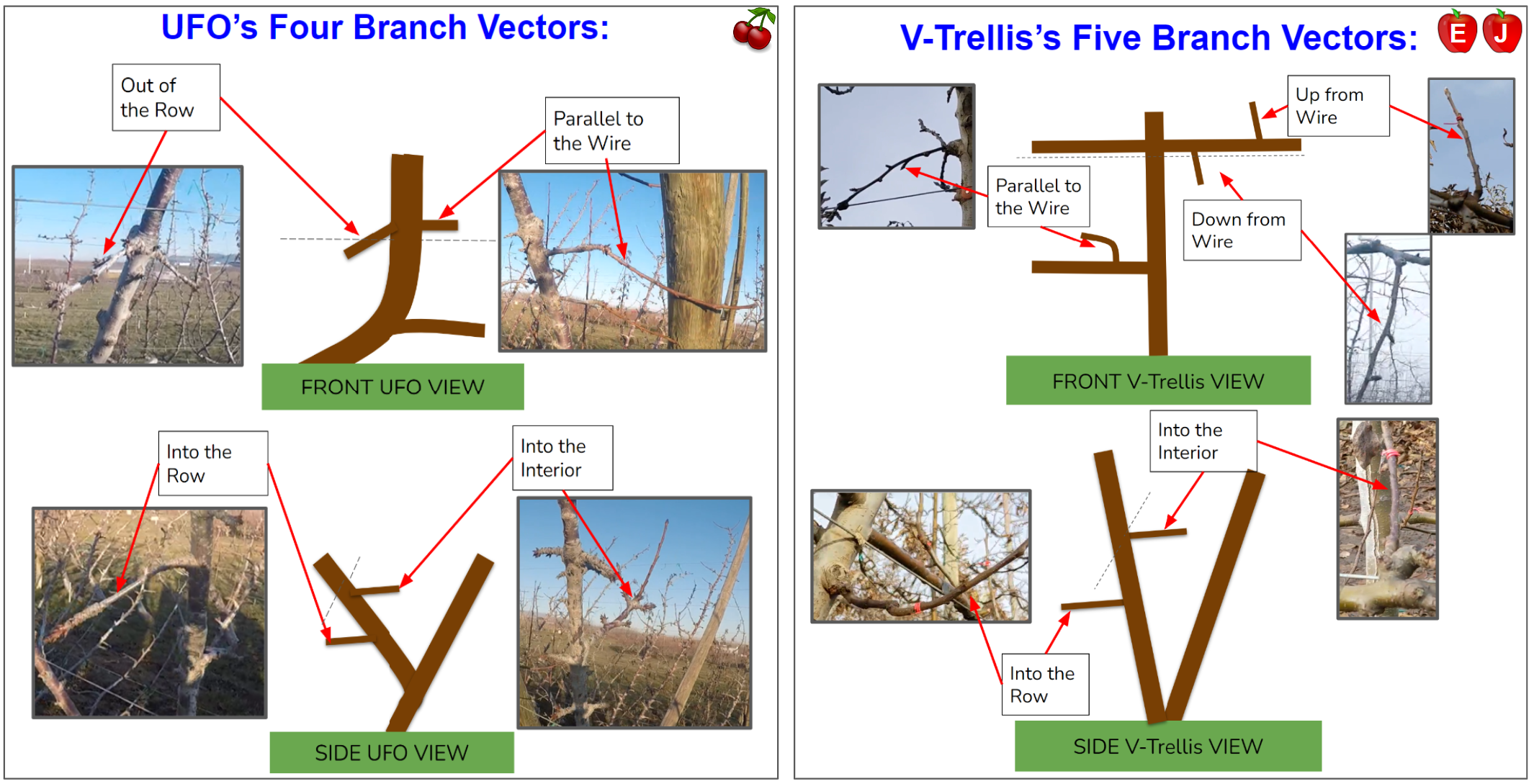}
    \caption{Tertiary branches refer to the small branches that produce fruit and leaves. These branches grow in different directions relative to the plane defined by the support posts and wires. In addition to length, qualitative terms were used to define the direction of growth of the tertiary branch relative to this support structure. The UFO architecture utilized four different terms, distinguishing between in and out of the plane and being parallel to the wire. The V-trellis architecture extended these terms to more specifically indicate the growth direction in the plane (up or down).}
    \label{fig:branch_vector}
\end{figure}

The Upright Fruiting Offshoot (UFO) for Bing Cherries have four main qualitative branch vectors (see Figure \ref{fig:branch_vector} ``UFO''): ``Into the Interior,''  ``Into the Row,'' ``Parallel to the Wire,'' and ``Out of the Row.'' ``Into the Interior'' and ``Into the Row'' are branches growing into or out of the plane defined by the UFO's wire structure. ``Parallel to the Wire'' are branches growing in the plane and parallel to the (horizontal) wires. Lastly, any other branches that do not satisfy other vector classifications are ``Out of the Row.'' 

The V-Trellis Architecture used on the Envy and Jazz Apples have five main branch vectors (see Figure \ref{fig:branch_vector} ``V-Trellis''): ``Into the Interior'', ``Into the Row'', ``Parallel to the Wire'', ``Up from the Wire'', and ``Down from Wire''. The first three terms have the same descriptions as above. ``Up from Wire'' and ``Down from Wire'' grow parallel to the planar tree architecture and wire, but point up (or down).

\begin{figure}
    \centering
    \includegraphics[width=0.95\linewidth]{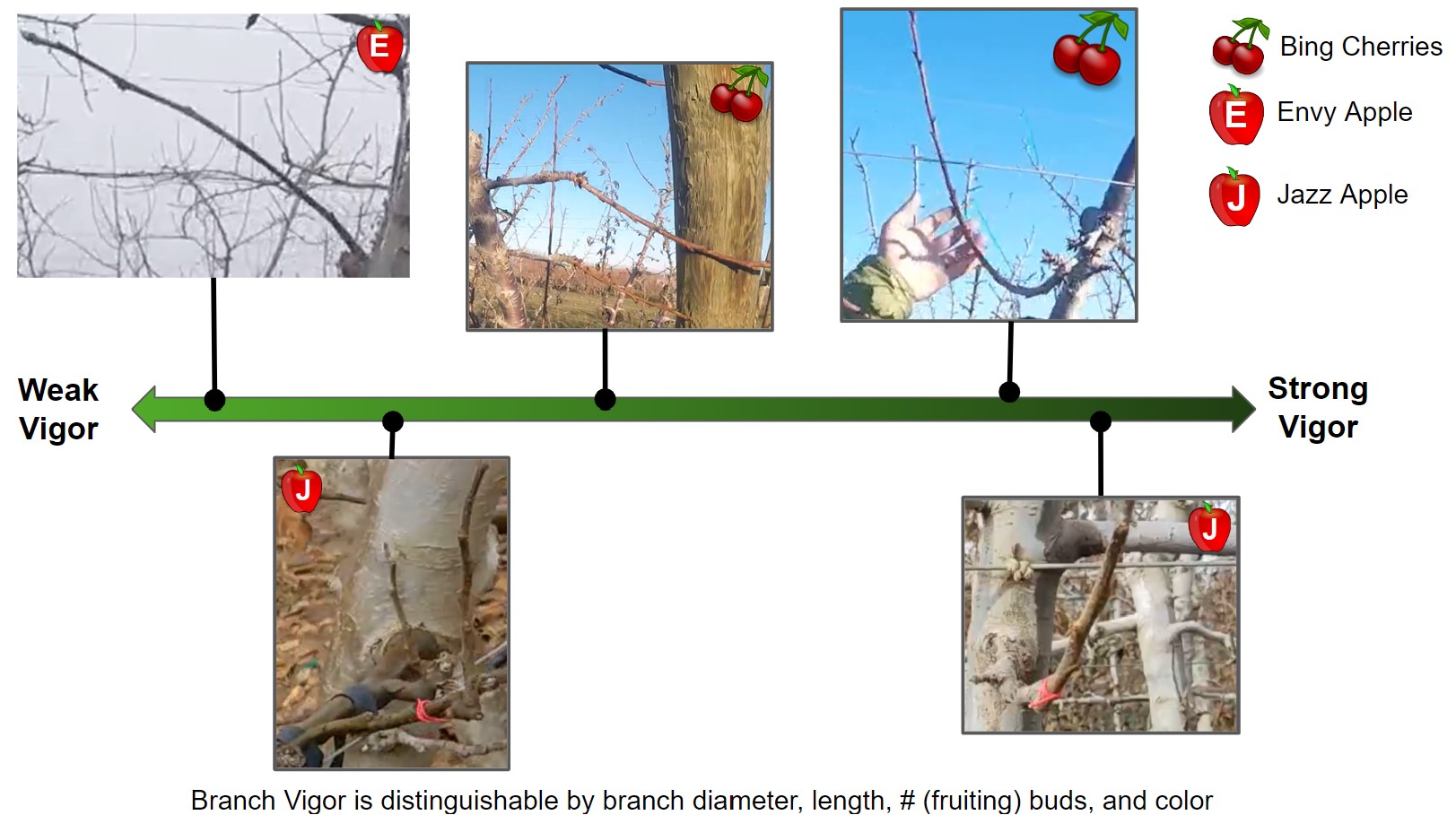}
    \caption{The term Vigor refers to how vigorous a tertiary branch is growing and is represented by a scale from weak to strong. Determining where a branch lies on the scale is a complex combination of length, diameter relative to the secondary branch, bud spacing and (sometimes) direction of growth. For example, Bing Cherry tertiary branches with \textbf{strong vigor} are often thicker, longer, have widely spaced buds, are shiny, and vertically. In contrast, \textbf{weak vigor} branches are shorter, thinner, and  droop.}
    \label{fig:branch_vigor}
\end{figure}

\textbf{Branch Vigor [General]} is a qualitative term that ranks the branch's growth on a scale from ``Weak'' to ``Strong''. Although somewhat subjective, vigor is determined by i) comparing the thickness of the tertiary branch to the support branch it grows out of, ii) the spacing of the buds on the branch, iii) the length of the branch, and iv) the color of the wood . The thicker the tertiary branch, the more vigor it has. Figure \ref{fig:branch_vigor} provides examples of both ``Strong'' and ``Weak'' vigor for all three tree cultivars.

\textbf{Type of (Most Common) Bud [Cherry]} refers to the ratio of fruiting versus vegetative bud on a tertiary branch. This term is a partial extension of the \textbf{Type of Branch} term as branches can possess both ``Fruiting Spur'' or ``Vegetative'' buds; a branch is designated a ``Fruiting spur'' or ``Vegetative'' depending on which bud type the branch has more of.  

\textbf{Number of Fruiting Spurs Before/After Cut [Cherry]} and \textbf{Number of Buds Before/After Cut [Apple]} terms refer to the estimated number of (fruiting spur) bud sites that exist before or after pruning. This is a numeric count of the number of buds along a branch (usually a tertiary branch). In the Bing cherry, fruiting buds (versus vegetative) are easily distinguished while in apples this is not always the case (and buds produce both leaves and fruit). 

\textbf{Observed Wire Number [Apple]} refers to which wire is closest to the pruned branch. Observed wire numbers start at 1 at the bottom and increase as they go up; a typical V-trellis has 5-7 wires spaced 18 inches apart. Because

%
%

\subsubsection{Pruning Explanations [Global/Local]} This category covers explanations for why a particular pruning cut was made (see Table~\ref{tab:prune_explanations}). These explanations can further be divided into {\em Global} versus {\em Local} explanations. Each explanation also has an associated context (in the form of a verbal description) of when and how that decision was made. These explanations help connect specific pruning cuts to higher-level horticultural concepts. 

\begin{table}[!h]
\small
\centering
\begin{tabular}{ |p{3.5cm}|p{7cm}|p{2.5cm}| }
\hline
\rowcolor{CherryBlossomPink}
 \textbf{Label} & \textbf{Definition} & \textbf{Example Label} \\
 \hline 
 \textbf{Global vs. Local} & Identification of the influence level of the pruning factors for the specific cut & Gobal, Local \\
 \hline
 
 \textbf{Pruning Context and Explanation:} & Verbal description of pruning explanations or context of the cut & ``Growing into the interior''  \\
 \hline

 \end{tabular}
 \caption{Pruning Explanations [Global/Local] terminology consists of two terms relating to the explanations provided by participants}
\label{tab:prune_explanations}
\end{table}

\textbf{Global versus Local} refers to  whether the pruning decision was primarily influenced by ''Global'' or ''Local'' factors. ``Global'' factors refer to horticultural concepts, such as those defined in~\ref{sec:contexts}, while ``Local'' factors are tied to terms defined in the previous subsection, such as ``Branch Vector'' or ``Type of Most Common Bud''. 

\textbf{Pruning Context and Explanation:} refers to the verbal description of the context and/or explanation provided while pruning. These explanations can be either Global or Local. An example of an explanation is ``Growing into the interior''. More explanations and contexts are reported in Section~\ref{sec:contexts}.

%
%

\subsubsection{Physical Cuts}
This category refers to the actual demonstrated cuts. This category is further split into three terms that describe the cut itself (see Table~\ref{tab:physical_cut}).


\begin{table}[!h]
\small
\centering
\begin{tabular}{ |p{3.5cm}|p{6.5cm}|p{2.5cm}| }
\hline
\rowcolor{CherryBlossomPink}
 \textbf{Label} & \textbf{Definition} & \textbf{Example Label} \\
 \hline 
 \textbf{Type of Cut} & The specific cut the pruner used on the branch & Tipping, Thinning, Heading\\
 \hline
 
 \textbf{Angle of Cut} & The physical angle of the lopper when cutting the branch relative to the main branch & $45^{o}$, Parallel to main branch  \\
 \hline
 
 \textbf{Cut Placement} & Where the loppers were placed and the branch was cut (given in inches from base or tip of branch) & 3" from tip  \\
 \hline
 
 \end{tabular}
 \caption{The Physical Cuts terminology category represents cut actions performed by an individual during pruning. Our analysis revealed three labels: Type of Cut, Angle of Cut, and Cut Placement. 
}
\label{tab:physical_cut}
\end{table}

\begin{figure}
    \centering
    \includegraphics[width=0.8\columnwidth]{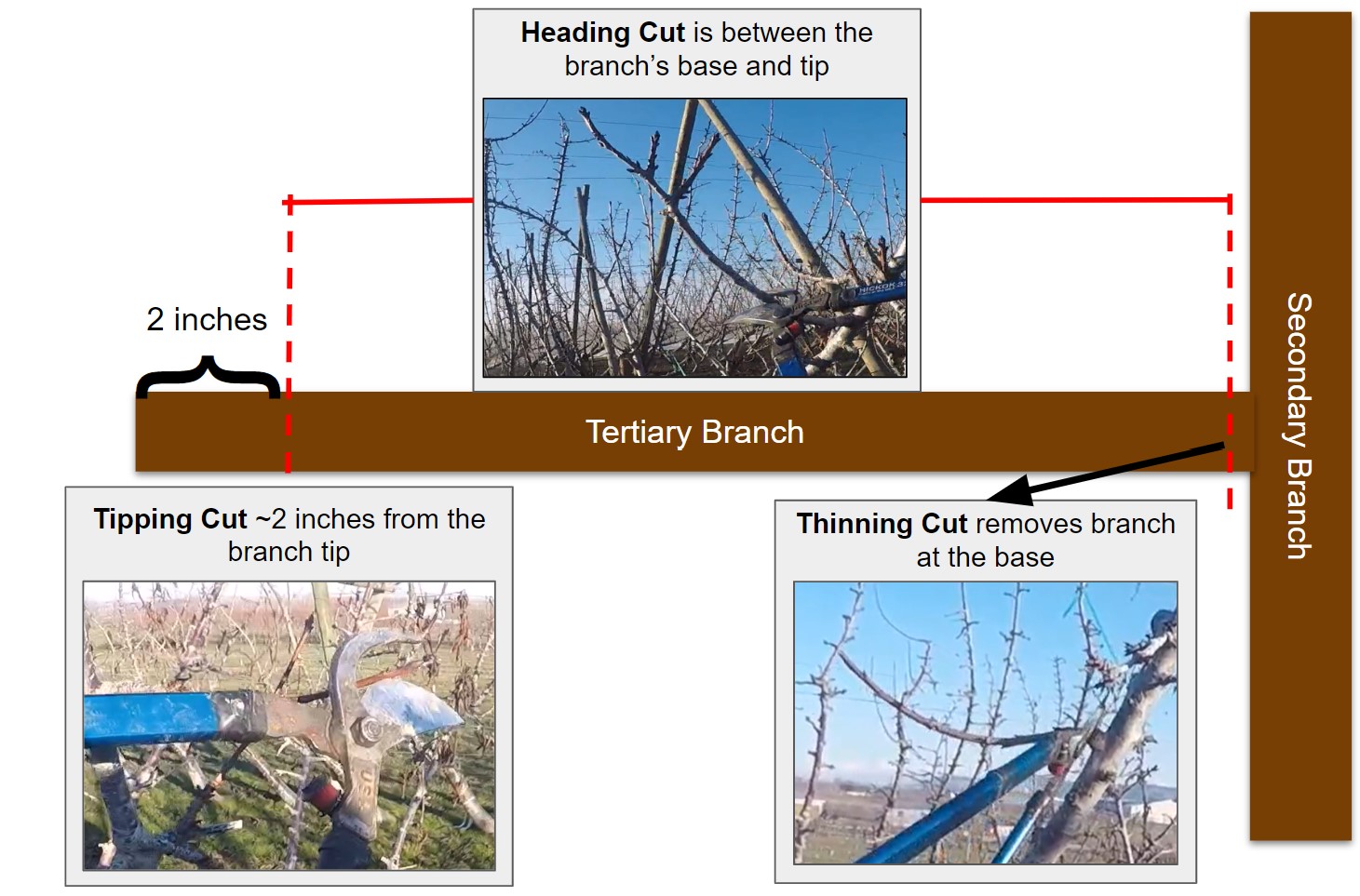}
    
    \caption{Observation of physical pruning cuts revealed three \textbf{Type of Cuts}: thinning, heading, and tipping. These cuts vary in the location where the tertiary branch is cut. For example, thinning cuts remove a branch at its base, while tipping cuts only remove the last two inches.}
    \label{fig:type_of_cuts}
\end{figure}

\textbf{Type of Cut} refers to the specific cut the participant used on the branch and is a general description of lopper placement on a branch (see Figure~\ref{fig:type_of_cuts}). We identified three main cut types: ``thinning'', ``heading'', and ``tipping''. These cuts differ by where they are made on the branch. A ``thinning cut'' removes the entire branch, with the loppers positioned any where from the base of the branch to approximately one to two inches out. In contrast, a ``tipping cut'' is placed at the end of the branch, and removes approximately two inches. Finally, a ``heading cut'' is placed somewhere between the two, i.e., from two to three inches from the base and the tip. 

\textbf{Angle of Cut} defines the physical angle of the lopper relative to the branch being cut. These are given as both descriptive and numerical terms. Descriptive includes relative placement, such as: ``Parallel to the main branch''.  Numerical refers to the angle of the loppers relative to the direction of the branch, the most common of which was ``45 degrees''.

\textbf{Cut Placement} refers to where the branch was cut as approximately measured in inches from either the base or the tip of the branch. This is related to the \textbf{Type of Cut} term, but is an (approximate) numerical measurement instead of a descriptive term.

\subsection{Results: Discovered Horticultural Contexts and Pruning Heuristics}
\label{sec:heuristics}

In this section we present the results of coding the horticultural contexts and heuristics uncovered in the analysis. These results build on the pruning terminology presented in the previous section. Our overall observation is that the horticulturists begin with the horticultural context for a cut and use that to describe why a cut was made, growers reference horticultural contexts when defining their pruning heuristics, but primarily focus on the heuristics, while the actual pruning is a visual and spatial task guided by those heuristics. 

We first present the horticultural contexts, along with the pruning heuristics that were related primarily to each context. We then present examples from actual pruning cuts that are representative of those heuristics, and which visual/spatial characteristics map to those cuts for each tree architecture and cultivars (UFO, Bing: Subsection~\ref{sec:bing}; V-trellis, Envy: Subsection~\ref{sec:envy}; V-trellis, Jazz: Subsection~\ref{sec:jazz}). Finally, we discuss the types of visual-spatial tasks pruners perform when pruning.

\begin{figure}[!h]
    \centering
    \includegraphics[width=\columnwidth, height=5.5cm]{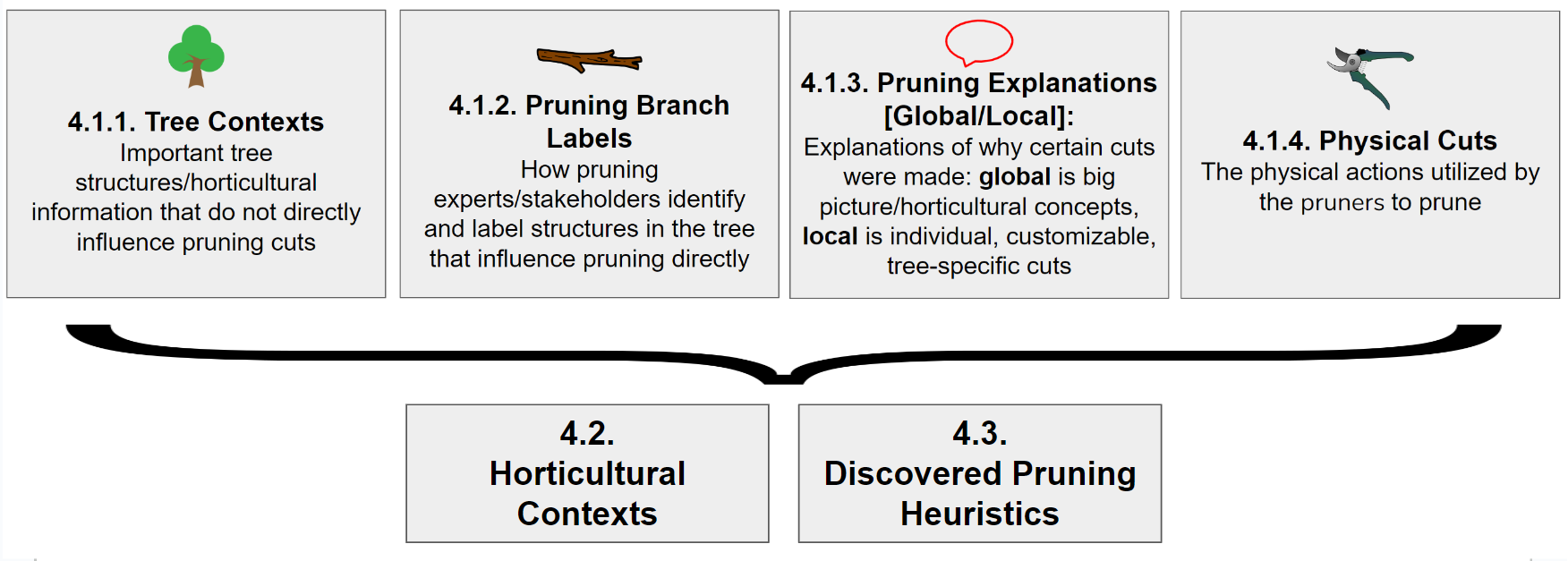}
    \caption{Summarized analysis technique mapping four terminology categories to horticultural contexts and discovered pruning heuristics}
    \label{fig:label_procedure}
\end{figure}

%
%

\subsubsection{Horticultural Contexts that Influence Pruning Decisions}
\label{sec:contexts}
\begin{figure}
    \centering
    \includegraphics[width=0.95\columnwidth]{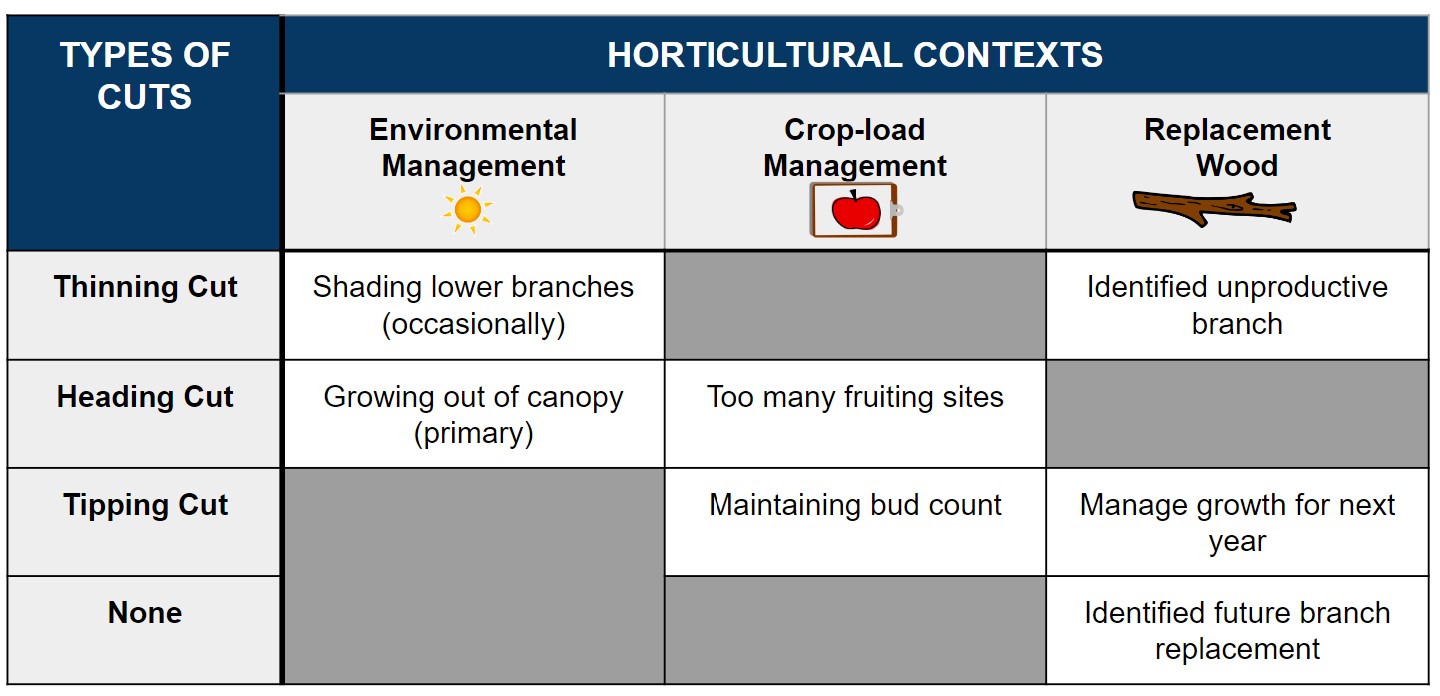}
    \caption{Pruning analysis revealed three main horticultural contexts that influence pruning decisions: environmental management, crop-load management, and replacement wood. Each context can map to at least one of the three cut types, i.e., thinning, heading, and tipping, alongside not cutting.}
    \label{fig:context_to_cut}
\end{figure}

Our grounded coding analysis of explanations provided during in-the-field interviews and pruning observations revealed three common horticultural motivations that influence pruning decisions (see Figure \ref{fig:context_to_cut}). These three contexts are: environment management, crop-load management, and replacement wood.

\subsubsection{Environmental Management} refers to pruning decisions that change the local environment. The primary environmental concern is sunlight (removing branches that shade others), but can also include air circulation (removing branches to improve circulation), fungus or bacterial growth (removing diseased branches). For example, ``... thought process is managing crop load, but also managing light in the block to thin it out, to allow for good channels of light from top to bottom''. Specifically, every piece of fruit needs to have good light and air circulation to be of high quality. An example of a pruning heuristic based on this is removing more branches at the top of tree to allow for more light to reach the fruit on the lower wires.

\subsubsection{Crop-load Management} refers to controlling the amount of (potential) fruit on a branch. For example, ``...the balance of quantity and quality of fruit''. Pruning (removing buds) is one of the methods used to control the quantity of fruit, which indirectly influences the quality, since more buds leads to more, but smaller, fruit~\cite{verma2022review,kon2018apple}. From a horticultural stand-point this is a complex calculation that considers the plant growth characteristics of the cultivar, the two-year growth cycle (buds for next year are set the previous year), and other effects (pollination, fertilization, thinning). Typical pruning heuristics based on this are 1) number of buds per branch and 2) even spacing of buds along the branch.


\subsubsection{Replacement Wood} refers to removing unproductive wood, which is characterized by wood with few buds with respect to the branch's length and diameter. Pruning for replacement wood involves two steps: 1) determining which branch to remove to encourage new growth, and 2) deciding which branch to leave to fill in space within the tree's architecture. Replacement wood pruning, therefore, considers current and future tree growth. For example, pruners will leave one or two pieces of lateral wood as possible secondary branch replacements to tie down on the tree architecture's wire system the following dormant pruning season after completely removing the current secondary branch.

%
%


\subsection{Upright Fruiting Offshoot Heuristics --- Bing Cherries:}
\label{sec:bing}
We report two heuristics for Bing Cherries growing with an Upright Fruiting Offshoot architecture for environmental management and crop-load management. These heuristics are derived 124 pruning cuts from five minutes of video data from one horticulturist. 

\subsubsection{Environmental Management}

\begin{figure}[!ht]
    \centering
    \includegraphics[width=\linewidth]{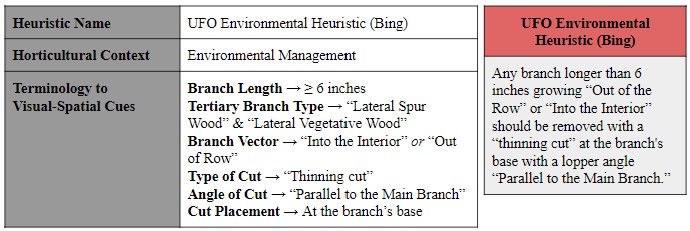}
    \label{fig:bing_environment}
\end{figure}




Analysis of the estimated \textbf{Branch Length} for Bing Cherries revealed the average estimated length of ``lateral'' branches pruned by the participant was 7.46 inches. Further investigation by \textbf{Tertiary Branch Type} showed the average length for ``Lateral Spur Wood'' at 6.93 inches and 9.39 inches for ''Lateral Vegetative Wood''. Out of the 124 cuts performed, we analyzed 106 branches pruned were longer than six inches, while only 18 cuts, or 14.5\% of all Bing Cherry pruning cuts, violated this rule. Most length violations occurred in ``Lateral Spur Wood'', with 12 cuts, and the remaining six for ``Lateral Vegetative Wood''. 

The horticulturist interview also explicitly identified branches with a specific \textbf{Branch Vector} as a pruning heuristic for Bing Cherries. Specifically, the horticulturist cited removing interior growth to prevent shading (environmental management) of lower branches. Recall that Bing Cherries --- due to the Upright Fruiting Offshoot architecture --- has four possible vectors (see Figure~\ref{fig:branch_vector}). Of the 124 cuts categorized, 77 were ``Parallel to the Wire'', 27 were ``Into the Row'', 8 ``Out of Row'', and 12 ``Into the Interior''. All 12 branches growing ``Into the Interior'' were removed with a ``Thinning Cut'' at the base of the branch and with an \textbf{Angle of Cut} ``Parallel to the Main Branch.'' Similarly, our analysis of ``Out of Row'' branches to \textbf{Type of Cut} also saw all 8 branches removed with ``Thinning Cuts'' at the base of the branch with the same lopper angle. 



\subsubsection{Crop-Load Management}


\begin{figure}[!ht]
    \centering
    \includegraphics[width=\linewidth]{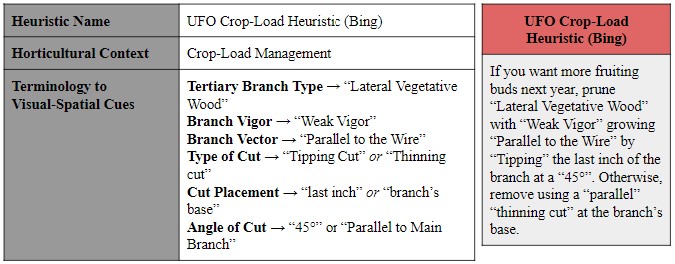}
    \label{fig:bing_cropload}
\end{figure}

Of the 64 ``Lateral Vegetative Wood” classified, we coded 36 unique pruning cuts (56.25\%) with a \textbf{Branch Vector} “Parallel to the wire.” 21 of these had a \textbf{Branch Vigor} considered “Weak” given shorter length to diameter ratio. For 16 of the 21 cuts that were ``Weak, Lateral Vegetative Wood'' growing ``Parallel to the Wire,'' the participant utilized a ``Thinning cut'' while the remaining five cuts were coded as ``Tipping'' (3) or ``Heading'' (2). All “Thinning cuts” had an angle ``Parallel to the Main Branch”. In contrast, the participant used a “45 degree” angle cut for all three tipping cuts and one of the two heading cuts. 

When the participant used a tipping or heading cut they justified it with the context of keeping or influencing future fruiting spur bud growth. Specifically, the horticulturist said tipping ``Lateral Vegetative Wood'' would result in the setting up fruiting spurs for following growing season. The horticulturist called this tactic as being ``greedy'' given the expectation that more fruiting spurs will set up for the next pruning season. Overall, tipping should be used by the pruner when growers or pruners desire more fruit for the next season, otherwise they should remove this ``Weak, Lateral Vegetative Wood'' growing ``Parallel to the Wire'' with a thinning cut.




\subsection{V-Trellis Heuristics --- Envy Apples:}
\label{sec:envy}
In this subsection, we report the heuristics for environmental management and crop-load management for Envy Apples with a V-trellis architecture. We derive these heuristics from quantitative analysis of 229 coded cuts from six minutes of video and three participants (two pruners and one grower). 

%
%

\subsubsection{Environmental Management:} 




\begin{figure}[!ht]
    \centering
    \includegraphics[width=\linewidth]{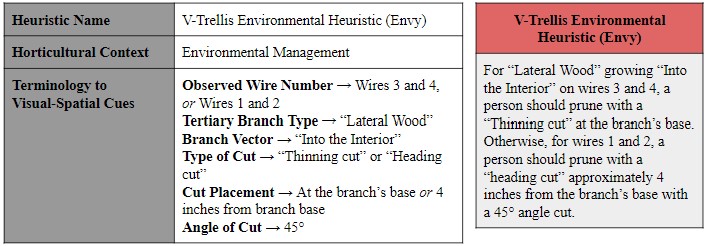}
    \label{fig:envy_environmental}
\end{figure}

Similar to Bing Cherries, the horticulturists cited pruning back branches growing ``Into the Interior'' for environmental management. From our analysis of 229 pruning cuts, we observed 36.24\% of these cuts (83 cuts) were ``Into the Interior.'' This \textbf{Branch Vector} was the most prominent of all five possible branch vectors: ``Into the Interior'', ``Parallel to the Wire,'' ``Up from Wire,'' ``Down from Wire,'' and ``Into the Row.'' Of these 83 cuts, 73 were ``Lateral Wood'' while the remaining 10 were ``Spurs.'' These 73 pruning cuts were distributed across different wires in the Envy Apple architecture. Recall that the Envy Apple tree utilized a 7-wire V-trellis architecture. Our pruning observations focused on the bottom four wires with wire \#1 being closest to the ground. For the 73 cuts of ``Lateral Wood'' growing ``Into the Interior'' 10 were on wire \#1, 6 on wire \#2, 34 on wire \# 3, and 20 on wire \#4.  


\begin{figure}
    \centering
    \includegraphics[width=0.95\linewidth]{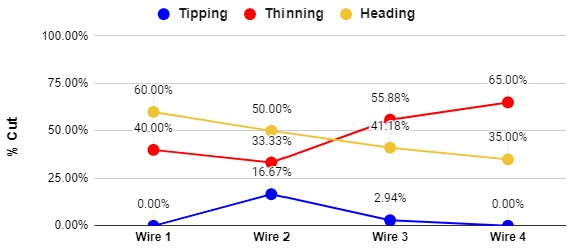}
    \caption{Analysis of the \textbf{Type of Cut} for ``Lateral Wood'' growing ``Into the Interior'' compared to \textbf{Observed Wire Number} saw an inverse relationship between ``heading'' and ``thinning'' cuts. Participants used ``thinning cuts'' on wires higher from the ground, and ``heading cuts'' otherwise to allow for more light to reach the lower wires.} 
    \label{fig:cut_vs_wire}
\end{figure}

Using this wire distribution, we analyzed what \textbf{Type of Cut} a participant used when pruning was dependent on where in the tree they were pruning. The grower explained during the in-field interview that there is a disparity in light distribution throughout the tree with the tops of a tree receiving more light than the bottom. As a result, pruners are more ``aggressive'' (i.e., remove more branches) when pruning on higher wires to help create channels for sunlight to reach lower branches. We observed an inverse linear relationship when analyzing \textbf{Type of Cut} versus \textbf{Observed Wire Number} between the heading cuts and thinning cuts as shown in Figure \ref{fig:cut_vs_wire}. The higher the wire from the ground, the more a pruner used a ``thinning cut,'' while for wires closer to the ground, the more a ``heading cut'' was used. For example, on wire \#4, 65\% of all the \textit{Type of Cuts} (13 of the 20 cuts) were ``thinning cuts'' compared to wire \#1 with 40\% (4 of the 10 cuts).


%
%
\subsubsection{Crop-Load Management:}

Apples, unlike Bing Cherries, need enough space between bud sites to allow for fruit to grow to a desired quality. During the in-field interviews, the grower described leaving enough space for an apple to grow. Likewise, the horticulturist quantitatively defined leaving approximately four inches or a ``fist length'' between each bud to produce the correct quality and quantity of apple. Any buds closer than this are removed.



\begin{figure}[!ht]
    \centering
    \includegraphics[width=\linewidth]{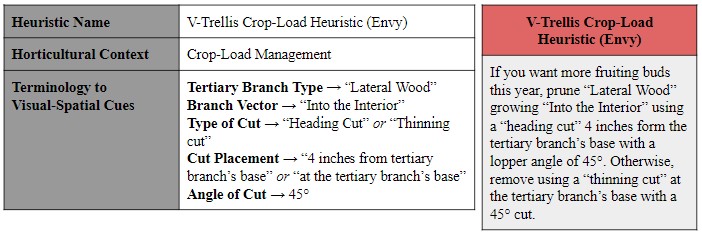}
    \label{fig:envy_cropload}
\end{figure}

We coded 194 ``Lateral Wood'' tertiary branches. 59 or 30.41\% of these branches had a \textbf{Branch Vector} ``Into the Row.'' Of these branches, participants pruned 54\% of all ``Into the Row'' growth with a ``heading cut'' approximately four inches from the base of the branch. The remaining 45.76\% of all cuts were thinning cuts. This fairly even distribution of ``heading'' and ``thinning'' cuts remained when accommodating for the \textbf{Branch Vigor}. 31 ``Into the Row'' branches had ``Strong'' vigor, and 28 had ``Weak'' vigor. Of the 28 ``Weak'' branches, there was an even (50\%) split between ``heading'' and ``thinning'' cuts with 14 branches for the two \textbf{Type of Cuts}. Likewise, ``heading'' cuts were more prominent in ``Strong'' vigor branches at 18 and 13 ``thinning'' cuts. Furthermore, all but one of the 32 ``heading'' cuts utilized an \textbf{Angle of Cut} of ``45 degrees.'' This \textbf{Angle of Cut} was preferred by the grower as it controls the angle of the new growth. 



By utilizing the ``heading'' cut compared to a ``thinning'' cut, we estimated 47 of the 197 buds (23.86\%) remained after participants pruned back ``Lateral Wood'' growing ``Into the Row''. The total percentage of buds kept is greater than the percentage of buds kept for all ``Lateral Wood'' at 18.28\%, or 104 buds out of 569. This cut behavior also decreased the average number of buds per branch from three to one.  

\subsection{V-Trellis Heuristics --- Jazz Apples:}
\label{sec:jazz}
In this subsection, we report three heuristics for Jazz Apples with a V-trellis architecture for all three horticultural contexts. We labeled a total of 308 from 2 minutes for four colored tags: 171 red, 129 green, 6 orange, and 2 white.

%
%

\subsubsection{Environmental Management:}
We extracted this heuristic from analysis of red tagged branches. The red tags marked prunes to keep the fruiting structure within a desired area around the secondary branch. 

\begin{figure}[!ht]
    \centering
    \includegraphics[width=\linewidth]{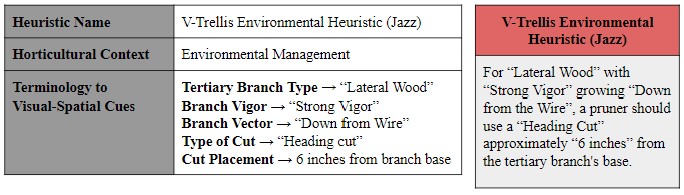}
    \label{fig:jazz_environment}
\end{figure}

From the informal grower interview, the desired \textbf{Cut Placement} --- represented by tag placement --- is anywhere between six and eight inches depending on where in the tree (what wire) the branch is growing from. The grower explained that the six inches was to stop the branch and apples from shading lower branches. We coded 171 branches with a red tag. The \textbf{Tertiary Branch Type} of all 171 branches was ``Lateral Wood,'' with 113 (66.1\%) having ``Strong'' \textbf{Branch Vigor} and 58 (33.9\%) being ``Weak''. Of these ``Lateral Wood'' branches, 28 had a \textbf{Branch Vector} ``Parallel to the Wire,'' 54 were ``Into the Row,'' 47 were ``Up from Wire,'' 37 were ``Into the Interior,'' and 5 ``Down from Wire.'' 



Using the red tag's placement along the branch, approximately 81\% of all \textbf{Type of Cuts} tagged by the grower and expert pruner reflected a ``heading cut,'' 16.3\% were ``thinning cuts,'' and 2.9\% were ``tipping cuts.'' Of these heading cuts, we see an average \textbf{Cut Placement} between four and six inches from the branch base. However, there were instances on the first wire from the ground --- the grower referred to this as the bottom layer --- when the tag placement was approximately eight inches away from the branch's base. The grower further explained the longer cut placement away from the secondary branch is fine on lower wires as there is less branches to block.

%
%
\subsubsection{Crop-Load Management:}
This heuristic is from our analysis of branches marked with a green tag. The green tag indicated areas to prune for spacing between buds for fruit to grow. 

\begin{figure}[!ht]
    \centering
    \includegraphics[width=\linewidth]{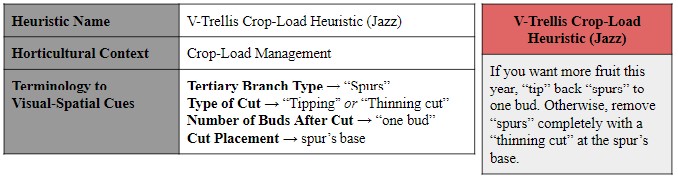}
    \label{fig:jazz_cropload}
\end{figure}

\begin{figure}
    \centering
    \includegraphics[width=0.95\linewidth]{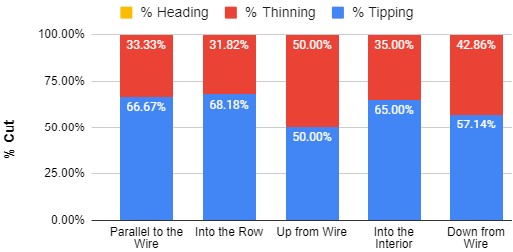}
    \caption{The distribution of \textbf{Type of Cut} vs. \textbf{Branch Vector} for ``spurs'' marked with a green tag in Jazz Apples. The ``tipping'' cut was utilized at least half of all cuts regardless \textbf{Branch Vector} the ``spur''. This prune allowed the grower to prune for spacing and keep more buds this harvest year. The only other cut marked were ``thinning'' cuts.}
    \label{fig:spur_cut}
\end{figure}

Throughout our categorization of tagged branches from the informal grower interview, we identified 129 green tags --- 84 being ``spur'', 45 being ``Lateral Wood'' --- containing approximately 280 buds. 160 of these buds belonged to ``Spurs'' averaging around 2 buds a spur. These spurs were tagged by the grower and expert pruner to remove with a ``tipping'' cut (64.29\% or 54 cuts) compared to ``thinning'' cuts (35.71\% or 30 cuts) and no ``heading'' cuts. This distribution of \textbf{Type of Cut} is consistent for ``spurs'' growing regardless of \textbf{Branch Vector}, as shown in Figure \ref{fig:spur_cut}.

One reason pruners tip buds back is it allows growers to keep enough buds for the current harvest year to produce fruit by leaving enough space for fruit to grow. Like Envy apples, Jazz apples need space between buds to allow fruit to grow to the desired quality. This spacing was identified by the grower during the informal grower interview to be three inches apart. 


By tagging ``spurs'' with a ``tipping'' cut, we estimated 53 of the 160 buds were kept across the tagged tree instances we analyzed, or about 33.1\% of all ``spur'' buds. This is higher than the percentage of buds kept for all \textbf{Tertiary Branch Types} at 28.21\%, or 79 of the 280 buds.

%
%

\subsubsection{Replacement Wood:}

These branches for replacement wood removal were tagged with a white flag. Furthermore, this heuristic was an explicitly stated rule by the grower during the informal interview. Namely, any secondary branch that was larger than 1.2 inches in diameter is pruned for replacement. Specifically, the white flag indicated the secondary branch was considered too vigorous. This vigor, as described by the grower, affects first-year growth and are considered less profitable due to producing less buds and having more erratic growth.  

\begin{figure}[!ht]
    \centering
    \includegraphics[width=\linewidth]{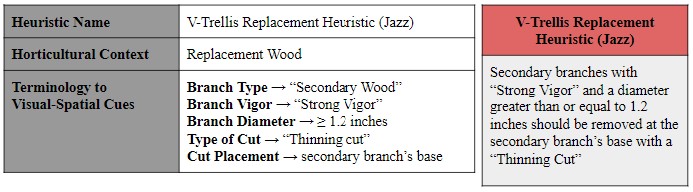}
    \label{fig:jazz_replacement}
\end{figure}


We coded two instances of branches tagged with a white flag. The placement of the flag was placed as close as the base of the secondary branch to six inches away. Furthermore, each white flag was also paired with an orange dot. Recall the orange dot signifies an area (about 6 inches around the dot) that pruners should \textit{not} prune. This action allows for new branches to grow that can be the ``replacement,'' or the next branch that will be wired down to fill space in the V-trellis architecture.

\subsection{Visual-Spatial Tasks Pruners Do}
\label{sec:tasks}

This subsection describes four visual-spatial decision tasks pruners perform while pruning: 

\begin{enumerate}
    \item Which bud to remove out of a bud cluster?
    \item Which branch do pruners leave for replacement?
    \item How far back to prune?
    \item How to identify extra vigor?
\end{enumerate}

These visual-spatial tasks are the in-the-field (local) pruning decisions that affect \textit{how} a pruner prunes. Each of these factors can be affected by more than one horticultural context defined in \ref{sec:contexts}.

\subsubsection{Which bud to remove out of a bud cluster?} In our example crop-load management heuristic for Jazz Apples, we describe removing all but one bud in a bud cluster. This decision allows growers to keep more buds while allowing each bud to have space for fruit to grow to their desired size and quality. This requires the pruner to decide which of the buds to remove and which to keep while pruning. 

Currently, we do not know what factors influences a pruner's decision to keep certain buds. Some possible factors could be the direction and vigor of the bud. Environmental management can also be a factor when considering how, if the bud produced a fruit, would shade the tree. Future exploration with pruning stakeholders is required to determine what influences this bud removal decision. 

\subsubsection{Which branch do pruners leave for replacement?} Of the three horticultural contexts that influence pruning, one was referred to as replacement wood. Replacement wood involves two parts: 1) completely removing an unproductive branch, and 2) replacing the branch with new wood for future growth. We identified with white flags in Jazz Apples when pruners perform the first step in replacement wood when a branch reaches a diameter greater than 1.2 inches. We also identified the area around the branch pruners do not prune to produce replacement wood (orange dots). However, we did not record any instances of branches that were used as replacement wood. 

One influential factor in pruner's selecting replacement wood is the ability of the branch to be wired down. For example, in Envy and Jazz Apples, replacement wood often replaces secondary branches along the wire system. Branches that growth with a \textbf{Branch Vector} ``Parallel to the Wire'' would be easier to tie down. Other factors such as \textbf{Branch Length} or \textbf{Branch Vigor} may also influence this decision.

\subsubsection{How far back to prune?} When pruning for sunlight management (light distribution) throughout a tree, pruners need to consider how far back to prune branches. This is also referred to pruning rigor by pruning stakeholders. 

In our analysis we learned that pruning rigor is influenced by where in the tree and how vigorous the branch. For example, ``strong'' vigor branches in Jazz Apples should be pruned more rigorously compared to ``weak'' vigor so branches to control \textbf{Branch Vigor}. Furthermore, the distance is also attributed to experimenting with pruning rigor and how it affects crop-load management over various seasons.


\subsubsection{How to identify extra vigor?} Recall that branches exist on a \textbf{Branch Vigor} scale with one end being ``weak'' and the other being ``strong.'' Pruners must be able to identify where along this scale a branch fits when pruning. They also need to identify when a branch is too vigorous. When branches become too vigorous, they impact fruit quality by stealing more tree resources compared to other tertiary branches in the tree and need to be completely removed. 

In this paper we presented example branches with different \textbf{Branch Vigor} for all three cultivars and the factors pruning stakeholders use to identify vigor (see Figure \ref{fig:branch_vigor}). We do not have definitive values of when growers identify branches as having too much vigor. Learning these values requires future studies with pruning stakeholders.

\subsection{Pruning Heuristics Discussion}
\label{sec:pruning_discuss}
In this section we presented the results of our grounded coding pruning heuristic analysis across 13 minutes of video from two pruning protocols (silent and speak-aloud) and tagged trees. We reported three contexts that influence pruning decisions and report seven heuristics mapping between high-level horticultural contexts to individual cuts for Bing Cherries, Envy Apples, and Jazz Apples. More pruning contexts may exist that are not presented in this paper. To learn these other contexts, this would require further investigation and interviews with the three pruning stakeholders. 

We also qualitatively introduced four visual-spatial tasks that affect \textit{how} a pruner prunes while in the field. These tasks were 1) which bud to remove in a cluster, 2) which branch do pruners leave for replacement wood, 3) how far back to prune, and 4) how to identify extra vigor. One area of future work is performing studies to quantify these visual-spatial tasks. Researchers can then implement these quantities into future autonomous pruning systems to more accurately identify these tree features and improve the system's pruning decision making.      

\section{Discussion}
\label{sec:discussion}

In this work, we designed and implemented three studies with six pruning stakeholders across three cultivars and two tree architectures. The first study used a semi-structured interview format, based on a critical decision analysis approach, to extract key pruning terminology and associated decision-making variables grounded in actual tree feature exemplars (e.g., branch geometry, buds). The second study represents a near-ideal version of the first study where every cut on 28 trees was labeled with the reason for the cut. Finally, we validated our pruning terminology and heuristics with a follow-up validation study with two horticulturists. 

Our grounded coding analysis allowed us to extract pruning terminology without {\em a priori} assumptions (section~\ref{sec:terminology}). Although other architectures and cultivars may have additional specific terms, the overall organization and terms of interest are likely to be conserved. One key observation, however, is that the terms {\em need to be grounded in physical exemplars}. In this study we relied on our participants providing exemplars in the orchard, e.g., pointing at branches that are ``too vigorous'' versus ones that are not. This relies on there being nearby exemplars that span the range of vigor in order to accurately capture what ``vigor'' means. In future work we propose using virtual tree models (where we can continuously change, e.g., the branch length) to more accurately capture how physical characteristics map to these qualitative terms. 

At a higher level, we extracted seven cultivar-specific pruning heuristics for three horticultural contexts --- environmental management, crop-load management, and replacement wood (section \ref{sec:heuristics}). The horticultural contexts are based in plant biology and are likely to be similar across all perennial crops. Moreover, they are also (relatively) easy to estimate from the physical plant geometry (e.g., modeling light transport (~\citet{boudon2012py}). Although the pruning heuristics are very cultivar-dependent, their overall structure (context plus physical features) is similar. These heuristics are also more amenable to automation, providing both a high-level goal (the context) plus what physical features (e.g., branch length, bud spacing) are used to make that decision. Again, the challenge is linking specific qualitative statements (``too far into the row'') to numerical quantities.

Numerous studies (and the growers themselves) have shown that the specific set of cuts a pruner chooses can vary widely, even on the same tree and even with the same pruner. While some cuts are unambiguous (e.g., removing excessively vigorous growth that shades lower branches) the majority of cuts are optimizing a function that has multiple solutions. If a branch has twelve buds, and should have at most ten, that is 12 choose 2 equals 66 combinations. While secondary considerations (bud spacing, chose buds on less vigorous branches) may reduce that number down, it is not surprising that you may see three or four different solutions. It is not clear if it is possible (or desirable) to capture this fine-grained decision making for purposes of automation.

We next provide observations and recommendations for future researchers interested in capturing pruning heuristics for autonomous pruning systems. 

\textbf{Going into the field is irreplaceable.} For determining what pruning heuristics exist based on a particular tree cultivar's architecture, we \textit{needed} to see example prunes in the field alongside the verbal explanations. This field interaction allowed our team to ask follow-up and what-if questions and for the interviewee to point to exemplars and visually demonstrate concepts. It also allowed us to capture a visual data set of various tree features that can be used in future autonomous vision systems.    

\textbf{Interacting with all pruning stakeholders is beneficial.} Through our study we identified three major stakeholders: horticulturists, growers, and pruners. Each stakeholder held different knowledge. If we had interacted with only the horticulturists we would have only uncovered the horticultural reasons for pruning, and not how those reasons were (on a practical level) implemented by the pruners as a visual-spatial task. 

\textbf{Putting numbers to terms and visual characteristics.} Bridging the gap between qualitative statements (``too vigorous'') and measurable quantities (number and spacing of buds, thickness and length of branch, texture and color of bark) is an open challenge. Being in the field and observing exemplars is a necessary first-step, but correctly mapping measurable quantities to these qualitative terms is likely to require a structured study for any complex quantity.

\textbf{How good should sensors be for successful pruning autonomous systems?} In order to implement the heuristics reported in this paper into physical autonomous pruning systems requires sensors. These sensors can measure how thick and long branch are and the spacing between buds. What is currently unknown is how accurate should these sensors be? Moreover, how much error can these sensors have that does not impact the robot's ability to make pruning decisions and perform cuts?

\section{Conclusion}

This paper presents the design and implementation of three studies across three tree cultivars --- Bing Cherries, Envy Apples, and Jazz Apples --- and two tree architectures --- Upright Fruiting Offshoot and V-Trellis. Our three studies were 1) a semi-structured interview with two pruning protocols (silent and speak-aloud), 2) an informal interview with tagged trees, and 3) an online validation interview. We analyzed six hours of video data of pruning observations and transcripts from six pruning stakeholders categorizing 661 distinct physical pruning cuts and tagged tree instances using a modified grounded coding analysis technique. This technique allowed us to discover a list of pruning terminology for communicating with stakeholders about pruning heuristics and a set of seven heuristics that connect global horticultural contexts to localized pruning decisions. This terminology and heuristics can be implemented into future autonomous pruning systems. We recommend that future researchers interested in autonomous pruning go out in the field and communicate with all three pruning stakeholders --- horticulturists, growers, and pruners --- to capture relevant pruning concepts and more holistic understanding of the decisions that influence pruning.

\section*{Statements and Declarations}

This material is based upon work supported by the AI Research Institutes program supported by NSF and USDA-NIFA under the AI Institute: Agricultural AI for Transforming Workforce and Decision Support (AgAID) award No, 2021-67021-35344

The first author is supported by the Achievement Rewards for College Scientists (ARCS) Foundation Scholar Oregon Chapter. 

There are no other conflicts of interest. 



\bmhead{Acknowledgements}
This material is based upon work supported by the AI Research Institutes program supported by NSF and USDA-NIFA under the AI Institute: Agricultural AI for Transforming Workforce and Decision Support (AgAID) award No, 2021-67021-35344

\bibliography{references}

\begin{thebibliography}{44}
\providecommand{\natexlab}[1]{#1}
\providecommand{\url}[1]{{#1}}
\providecommand{\urlprefix}{URL }
\providecommand{\doi}[1]{\url{https://doi.org/#1}}
\providecommand{\eprint}[2][]{\url{#2}}
 \bibcommenthead

\bibitem[{{Ag America Lending}(2022)}]{agamerica2022labor}
{Ag America Lending} (2022) (infographic) the u.s. labor shortage

\bibitem[{Akbar et~al(2016)Akbar, Elfiky, and Kak}]{akbar2016novel}
Akbar SA, Elfiky NM, Kak A (2016) A novel framework for modeling dormant apple trees using single depth image for robotic pruning application. In: 2016 IEEE international conference on robotics and automation (ICRA), IEEE, pp 5136--5142

\bibitem[{Alrigiwirsah et~al(2020)Alrigiwirsah, Lubis, and Novita}]{alridiwirsah2020effect}
Alrigiwirsah A, Lubis RM, Novita A (2020) The effect of pruning and chicken manure on vegetative growth of honey deli (syzygiumaqueum burn f.) in 9 months age. In: Proceeding International Conference Sustainable Agriculture and Natural Resources Management (ICoSAaNRM)

\bibitem[{Botterill et~al(2017)Botterill, Paulin, Green, Williams, Lin, Saxton, Mills, Chen, and Corbett-Davies}]{botterill2017robot}
Botterill T, Paulin S, Green R, et~al (2017) A robot system for pruning grape vines. Journal of Field Robotics 34(6):1100--1122

\bibitem[{Boudon et~al(2012)Boudon, Pradal, Cokelaer, Prusinkiewicz, and Godin}]{boudon2012py}
Boudon F, Pradal C, Cokelaer T, et~al (2012) L-py: an l-system simulation framework for modeling plant architecture development based on a dynamic language. Frontiers in plant science 3:76

\bibitem[{Bryant and Charmaz(2007)}]{bryant2007sage}
Bryant A, Charmaz K (2007) The Sage handbook of grounded theory. Sage

\bibitem[{Carroll(2017)}]{carroll2017annual}
Carroll B (2017) Annual pruning of fruit trees

\bibitem[{Charmaz(2014)}]{charmaz2014constructing}
Charmaz K (2014) Constructing grounded theory, 2nd edn. sage

\bibitem[{Chun and Knight(2020)}]{chun2020robot}
Chun B, Knight H (2020) The robot makers: An ethnography of anthropomorphism at a robotics company. ACM Transactions on Human-Robot Interaction (THRI) 9(3)

\bibitem[{Corbett-Davies et~al(2012)Corbett-Davies, Botterill, Green, and Saxton}]{corbett2012aivines}
Corbett-Davies S, Botterill T, Green R, et~al (2012) An expert system for automatically pruning vines. In: Proceedings of the 27th Conference on Image and Vision Computing New Zealand. Association for Computing Machinery, New York, NY, USA, IVCNZ '12, p 55–60, \doi{10.1145/2425836.2425849}

\bibitem[{Daniels(2018)}]{daniels2018strawberries}
Daniels J (2018) From strawberries to apples, a wave of agriculture robotics may ease the farm labor crunch

\bibitem[{Du et~al(2011)Du, Chen, Zhang, Scharf, and Whiting}]{du2011mechanical}
Du X, Chen D, Zhang Q, et~al (2011) Mechanical harvesting of ufo cherry: Investigation of tree plant dynamics. In: 2011 Louisville, Kentucky, August 7-10, 2011, American Society of Agricultural and Biological Engineers, p~1

\bibitem[{Economy(2019 [Online])}]{nae2019immigrants}
Economy NA (2019 [Online]) Immigrants and american agriculture

\bibitem[{Fallatah et~al(2020)Fallatah, Chun, Balali, and Knight}]{fallatah2020would}
Fallatah A, Chun B, Balali S, et~al (2020) 'would you please buy me a coffee?' how microcultures impact people's helpful actions toward robots. In: Proceedings of the 2020 ACM Designing Interactive Systems Conference

\bibitem[{Gallardo et~al(2012)Gallardo, Taylor, and Hinman}]{gallardo2012cost}
Gallardo RK, Taylor MR, Hinman HR (2012) 2009 cost estimates of establishing and producing Gala apples in Washington. Washington State University Extension

\bibitem[{He and Schupp(2018)}]{he2018sensing}
He L, Schupp J (2018) Sensing and automation in pruning of apple trees: A review. Agronomy 8(10):211

\bibitem[{Henwood and Pidgeon(2003)}]{henwood2003grounded}
Henwood K, Pidgeon N (2003) Grounded theory in psychological research.

\bibitem[{Kaba and Abunyewa(2021)}]{kaba2021new}
Kaba JS, Abunyewa AA (2021) New aboveground biomass and nitrogen yield in different ages of gliricidia (gliricidia sepium jacq.) trees under different pruning intensities in moist semi-deciduous forest zone of ghana. Agroforestry Systems 95(5):835--842

\bibitem[{Karkee et~al(2014)Karkee, Adhikari, Amatya, and Zhang}]{karkee2014identification}
Karkee M, Adhikari B, Amatya S, et~al (2014) Identification of pruning branches in tall spindle apple trees for automated pruning. Computers and Electronics in Agriculture 103:127--135

\bibitem[{Klein et~al(1989)Klein, Calderwood, and Macgregor}]{klein1989critical}
Klein GA, Calderwood R, Macgregor D (1989) Critical decision method for eliciting knowledge. IEEE Transactions on systems, man, and cybernetics 19(3):462--472

\bibitem[{Kolmani{\v{c}} et~al(2021)Kolmani{\v{c}}, Strnad, Kohek, Benes, Hirst, and {\v{Z}}alik}]{kolmanivc2021algorithm}
Kolmani{\v{c}} S, Strnad D, Kohek {\v{S}}, et~al (2021) An algorithm for automatic dormant tree pruning. Applied Soft Computing 99:106931

\bibitem[{Kon and Schupp(2018)}]{kon2018apple}
Kon TM, Schupp JR (2018) Apple crop load management with special focus on early thinning strategies: A usa perspective. Hort Rev 46:255--298

\bibitem[{Kupka(2007)}]{kupka2007growth}
Kupka I (2007) Growth reaction of young wild cherry (prunus avium l.) trees to pruning. Journal of Forest Science 53(12):555--560

\bibitem[{Lang et~al(2015)Lang, Musacchi, Whiting et~al}]{lang2015cherry}
Lang GA, Musacchi S, Whiting MD, et~al (2015) Cherry training systems

\bibitem[{Maughan et~al(2017)Maughan, Black, and Roper}]{maughanbackyardpruning}
Maughan T, Black B, Roper T (2017) Training and pruning apple trees. Accessed November 20, 2022 [Online]

\bibitem[{Mohajan and Mohajan(2022)}]{mohajan2022development}
Mohajan D, Mohajan H (2022) Development of grounded theory in social sciences: A qualitative approach

\bibitem[{Pandit(1996)}]{pandit1996creation}
Pandit NR (1996) The creation of theory: A recent application of the grounded theory method. The qualitative report 2(4):1--15

\bibitem[{Persello et~al(2019)Persello, Grechi, Boudon, and Normand}]{persello2019nature}
Persello S, Grechi I, Boudon F, et~al (2019) Nature abhors a vacuum: Deciphering the vegetative reaction of the mango tree to pruning. European Journal of Agronomy 104:85--96

\bibitem[{Polomski(2019)}]{polomski2019backyardpruning}
Polomski RF (2019) Pruning \& training apple \& pear trees. Accessed November 20, 2022 [Online]

\bibitem[{Poni et~al(2016)Poni, Tombesi, Palliotti, Ughini, and Gatti}]{poni2016mechanical}
Poni S, Tombesi S, Palliotti A, et~al (2016) Mechanical winter pruning of grapevine: Physiological bases and applications. Scientia Horticulturae 204:88--98

\bibitem[{Robinson et~al(2008)Robinson, Hoying, and Reginato}]{robinson2008tall}
Robinson T, Hoying S, Reginato G (2008) The tall spindle planting system: Principles and performance. In: IX International Symposium on Integrating Canopy, Rootstock and Environmental Physiology in Orchard Systems 903, pp 571--579

\bibitem[{Robinson et~al(2006)Robinson, Hoying, and Reginato}]{robinson2006tall}
Robinson TL, Hoying SA, Reginato GH (2006) The tall spindle apple production system. New York fruit quarterly 14(2):21--28

\bibitem[{Rothwell et~al(2022)Rothwell, Stone, and Davids}]{rothwell2022investigating}
Rothwell M, Stone J, Davids K (2022) Investigating the athlete-environment relationship in a form of life: An ethnographic study. Sport, Education and Society 27(1)

\bibitem[{Schupp et~al(2017{\natexlab{a}})Schupp, Winzeler, Kon, Marini, Baugher, Kime, and Schupp}]{schupp2017method}
Schupp JR, Winzeler HE, Kon TM, et~al (2017{\natexlab{a}}) A method for quantifying whole-tree pruning severity in mature tall spindle apple plantings. HortScience 52(9):1233--1240

\bibitem[{Schupp et~al(2017{\natexlab{b}})Schupp, Winzeler, Kon, Marini, Baugher, Kime, and Schupp}]{schupp2017pruningwholetreeseverity}
Schupp JR, Winzeler HE, Kon TM, et~al (2017{\natexlab{b}}) A method for quantifying whole-tree pruning severity in mature tall spindle apple plantings. HortScience horts 52(9):1233 -- 1240. \doi{10.21273/HORTSCI12158-17}

\bibitem[{Sevila(1985)}]{sevila1985robot}
Sevila F (1985) A robot to prune the grapevine. In: Agri-Mation 1, Chicago, Ill.(USA), 25-28 Feb 1985, ASAE

\bibitem[{Silverman et~al(2021)Silverman, Baroiller, and Hemer}]{silverman2021culture}
Silverman GS, Baroiller A, Hemer SR (2021) Culture and grief: Ethnographic perspectives on ritual, relationships and remembering

\bibitem[{Verma et~al(2022)Verma, Sharma, Sharma, and Chauhan}]{verma2022review}
Verma P, Sharma S, Sharma N, et~al (2022) Review on crop load management in apple (malus x domestica borkh.). The Journal of Horticultural Science and Biotechnology pp 1--23

\bibitem[{Von~Feigenblatt and Acu{\~n}a(2021)}]{von2021two}
Von~Feigenblatt O, Acu{\~n}a B (2021) Two case studies dealing with social studies at the secondary level: Applied anthropology and grounded theory. Journal of Alternative Perspectives in the Social Sciences 11(2):237--252

\bibitem[{Westling et~al(2021)Westling, Underwood, and Bryson}]{westling2021lidar}
Westling F, Underwood J, Bryson M (2021) A procedure for automated tree pruning suggestion using lidar scans of fruit trees. Computers and Electronics in Agriculture 187:106274. \doi{https://doi.org/10.1016/j.compag.2021.106274}

\bibitem[{Whiting et~al(2018)}]{whiting2018precision}
Whiting M, et~al (2018) Precision orchard systems. Automation in tree fruit production: Principles and practice pp 75--93

\bibitem[{Whiting(2011)}]{whiting2011ufo}
Whiting MD (2011) The ufo architecture: A novel system for high efficiency sweet cherry orchards. In: HORTSCIENCE, AMER SOC HORTICULTURAL SCIENCE 113 S WEST ST, STE 200, ALEXANDRIA, VA 22314~…, pp S124--S124

\bibitem[{You et~al(2022)You, Grimm, Silwal, and Davidson}]{you2022semantics}
You A, Grimm C, Silwal A, et~al (2022) Semantics-guided skeletonization of upright fruiting offshoot trees for robotic pruning. Computers and Electronics in Agriculture 192:106622

\bibitem[{Zhang et~al(2022)Zhang, Cai, Zhang, Li, Zhou, Chen, Mi, Jin, Xu, Yu et~al}]{zhang2022different}
Zhang D, Cai W, Zhang X, et~al (2022) Different pruning level effects on flowering period and chlorophyll fluorescence parameters of loropetalum chinense var. rubrum. PeerJ 10:e13406

\end{thebibliography}

\end{document}